\documentclass{article}

\PassOptionsToPackage{numbers, compress}{natbib}
\usepackage[preprint]{neurips_2026}
\usepackage{amsmath}
\usepackage{amssymb}
\usepackage{graphicx}

\usepackage[utf8]{inputenc} 
\usepackage[T1]{fontenc}    
\usepackage{hyperref}       
\usepackage{url}            
\usepackage{booktabs}       
\usepackage{amsfonts}       
\usepackage{nicefrac}       
\usepackage{microtype}      
\usepackage{xcolor}         

\title{WMAttack: Automated Attack Search for Adversarial Evaluation of World-Model Agents}

\author{%
  Zhixiang Guo \\
  Nanyang Technological University \\
  \texttt{zhixiang004@e.ntu.edu.sg}
  \And
  Siyuan Liang \\
  Nanyang Technological University \\
  \texttt{siyuan.liang@ntu.edu.sg}
  \And
  Shi Fu \\
  Nanyang Technological University \\
  \texttt{SHI011@e.ntu.edu.sg}
  \And
  Cheng Guo \\
  Nanyang Technological University \\
  \texttt{guo007@e.ntu.edu.sg}
  \And
  Andr\'as Balogh \\
  University of Szeged \\
  \texttt{abalogh@inf.u-szeged.hu}
  \And
  M\'ark Jelasity \\
  University of Szeged \\
  \texttt{jelasity@inf.u-szeged.hu}
  \And
  Dacheng Tao \\
  Nanyang Technological University \\
  \texttt{dacheng.tao@ntu.edu.sg}
}

\begin{document}

\maketitle

\begin{abstract} 
Despite the growing use of world models as decision-making agents, their adversarial robustness remains underexplored due to the lack of dedicated automated evaluation methods.  
A key obstacle is that attack evaluation must be both accurate and efficient: weak manually tuned attacks can overestimate robustness, while exhaustive hyperparameter search is prohibitively expensive because each candidate requires closed-loop rollouts through learned latent dynamics.  
We introduce \textbf{WMAttack}, an automated attack-search framework for adversarial evaluation of world-model agents.  
WMAttack formulates robustness evaluation as a finite-budget search over attack configurations, including attack families, perturbation budgets, optimization steps, restarts, and allocation rules.  
To improve search accuracy, \textbf{Self-Correcting Attack Search} (SCAS) refines the attack proposal distribution using feedback from reward degradation, action instability, runtime cost, and rollout variability.  
To improve search efficiency, \textbf{Representation-Guided Attack Retrieval} (RGAR) retrieves effective historical configurations from representation-similar tasks, providing a warm start for unseen environments.  
We provide a theoretical explanation showing that proposal refinement improves finite-budget search when it shifts probability mass toward high-utility attacks.  
Across Atari and DeepMind Control tasks, WMAttack consistently discovers stronger attacks than the evaluated baselines, improving normalized reward drop from $0.497$ to $1.034$ on DreamerV3 Atari and from $0.319$ to $0.682$ on DMC.
Ablations further show that RGAR improves initial candidate quality and SCAS improves final attack utility under fixed evaluation budgets. 
\end{abstract}

\section{Introduction}
World models have emerged as a foundational paradigm for decision-making agents, enabling sample-efficient planning and prediction through learned latent dynamics \citep{ha2018worldmodels,hafner2019planet,hafner2023dreamerv3}. 
Their growing use has also motivated safety-oriented world-model methods that incorporate constraints, uncertainty, or Lyapunov-style critics into model-based reinforcement learning~\citep{huang2024safedreamer,as2024safeexploration,zhang2024safedeepmbrl}. 
However, superior performance on visual control~\citep{tassa2018deepmind}, Atari~\citep{bellemare2013arcade}, and continuous-control benchmarks~\citep{hansen2024tdmpc2} does not guarantee robustness. 
Recent work on robust latent representations, latent safety filters, and surprise-based rejection further suggests that failures often arise from representation shift and out-of-distribution transitions~\citep{sun2024latentdynamic,seo2025unisafe,zollicoffer2025surprise}. 
Small, budget-constrained perturbations can propagate through learned latent representations and predictive dynamics~\citep{micheli2023iris,guo2025copyrightshield,guo2026physcondwma}, latent transition models~\citep{zhang2024storm}, policies~\citep{zhang2020samdp}, or adversarially corrupted planning objectives and trajectory rankings~\citep{ye2024robustmbrl,kobayashi2025lira,radji2025confuse,duan2026trap}, leading to unstable decisions or severe performance degradation. 
Despite these risks, we still lack a dedicated automated attack-search framework~\citep{liu2025metadv} to systematically evaluate world-model agents across diverse tasks, models, and attack families. 
This raises a critical question: how can we efficiently expose and quantify the failure modes of world-model agents under adversarial conditions?

Evaluating robustness in this context is challenging because attack efficacy is highly sensitive to both environment dynamics and adversarial hyperparameters~\citep{yang2024unisim,hu2023gaia,panfilov2026claudini}. 
World-model environments differ in observation spaces, reward scales, temporal structures, and policy sensitivities~\citep{hansen2024tdmpc2,vafa2024evaluating}, while robustness itself involves return degradation, action instability, and latent representation drift~\citep{zhang2020samdp,upadhyay2026worldbench,balogh2026verification}. 
Manual tuning is therefore unreliable and inefficient: weak attacks may create pseudo-robustness, while exhaustive search is computationally prohibitive because each candidate may require closed-loop rollouts and repeated latent-dynamics computation~\citep{croce2020autoattack,croce2021robustbench,liang2022large}. 
An effective evaluation method must balance accuracy and efficiency by avoiding cold-start search and adaptively refining configurations from agent feedback. 
These challenges motivate an automated framework that leverages cross-task experience and concentrates limited evaluation budgets on high-impact adversarial regions.
\begin{figure*}[t]
    \centering
    \includegraphics[width=\textwidth]{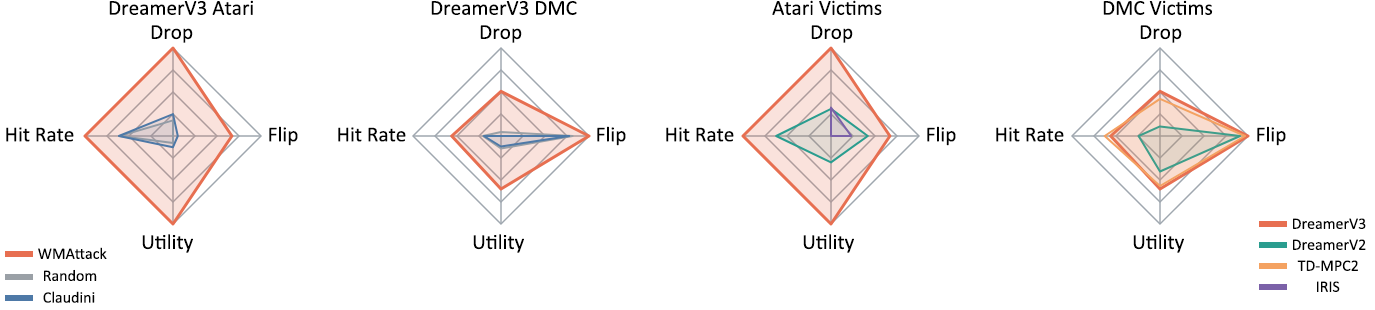}
    \caption{
    Overview of WMAttack effectiveness across tasks and world-model agents.
    Metrics are min-max normalized for visualization only; larger radius indicates stronger attack or search outcome, with exact values reported in Tables~\ref{tab:main-results} and~\ref{tab:efficiency-threshold}.
    }
    \label{fig:radar-summary}
\end{figure*}

In this work, we propose \textbf{WMAttack}, an automated attack-search framework for adversarial evaluation of world-model agents. 
Rather than relying on fixed attacks or heuristic manual tuning, WMAttack formulates robustness evaluation as a finite-budget search problem over adversarial configurations, including perturbation budgets, optimization steps, restarts, schedules, and allocation rules. 
To navigate this large search space, WMAttack employs a two-stage optimization process. 
First, \textbf{Representation-Guided Attack Retrieval (RGAR)} exploits intrinsic similarities between world models by projecting diverse dynamics into a unified latent representation space and retrieving effective historical attack configurations as a warm start for new tasks. 
This enables cross-task experience transfer and improves initialization over zero-knowledge starting points. 
Second, \textbf{Self-Correcting Attack Search} (SCAS) iteratively refines the attack strategy by using multi-dimensional feedback, such as reward degradation and action instability, as corrective signals to update a proposal distribution over the attack parameter space. 
By shifting probability mass toward regions of higher adversarial efficacy, WMAttack concentrates its search on potent configurations, reduces redundant trials, and provides a theoretical basis for discovering stronger attacks under a fixed evaluation budget.

Our contributions are summarized as follows:
\begin{itemize}
    \item We formulate adversarial robustness evaluation for world models as a finite-budget attack-configuration search problem, where the goal is to efficiently identify high-impact attacks over diverse tasks, models, attack methods, and hyperparameter settings.
    \item We develop \textbf{WMAttack}, an automated attack-search framework that combines \textbf{Representation-Guided Attack Retrieval (RGAR)} with \textbf{Self-Correcting Attack Search (SCAS)}, and provide a theoretical explanation for why feedback-driven proposal refinement can improve attack-search efficiency.
    \item We evaluate WMAttack on 46 DreamerV3 Atari and DeepMind Control tasks, where it discovers stronger attacks than random search and a Claudini-style baseline, and further report cross-model evaluations on DreamerV2, TD-MPC2, and IRIS, showing that it discovers stronger attacks than random search and a Claudini-style autoresearch baseline, with consistent gains in reward drop, action instability, and scalarized attack utility.
\end{itemize}

\section{Related Work}

\subsection{World Models in Reinforcement Learning}

World models are a central framework in model-based reinforcement learning, enabling agents to learn predictive latent dynamics for planning and decision making. 
Unlike vision-language-action models that often map multimodal inputs directly to actions through foundation models \citep{brohan2023rt}, RL-based world models emphasize environment dynamics, imagination, and sample-efficient control. 
Early methods such as PlaNet \citep{hafner2019planet} and Dreamer \citep{hafner2020dreamer} introduced recurrent state-space models for learning compact latent representations from pixels. 
Subsequent systems, including DreamerV2 and DreamerV3, scaled latent imagination to Atari, continuous control, and diverse RL domains \citep{hafner2021dreamerv2,hafner2023dreamerv3}. 
More recent architectures replace or augment recurrent dynamics with Transformer-based sequence modeling, as in IRIS and STORM \citep{micheli2023iris,zhang2024storm}, or generative transition models, as in DIAMOND \citep{alonso2024diamond}. 
Our work focuses on this class of RL-based world-model agents because their decisions are mediated by learned latent dynamics: perturbations to observations can propagate through the encoder, transition model, and policy, making closed-loop adversarial evaluation especially important~\citep{zhang2024visual,wang2025black,kong2025universal}.

\subsection{Adversarial Robustness and Evaluation}

Adversarial examples show that small, bounded perturbations can cause severe failures in neural models \citep{goodfellow2015explaining,wei2018transferable,liang2024object}. 
In reinforcement learning, the problem is more complex because perturbations affect actions, future states, and long-horizon returns. 
Prior work formalizes observation attacks through the State-Adversarial MDP framework \citep{zhang2020samdp} and shows that policies are vulnerable to observation perturbations, strategically timed attacks, and adversarial policies \citep{huang2017adversarial,lin2017tactics,gleave2020adversarial}. 
For world-model agents, these risks are amplified by learned latent dynamics~\citep{xu2026ctrlattack}: corrupted observations may distort latent states and imagined trajectories, leading to compounding errors in downstream control.

Robustness evaluation for world models remains fragmented. 
Recent diagnostic benchmarks such as WorldBench and iWorld-Bench study physical reasoning, interaction, or causal consistency in predictive models \citep{upadhyay2026worldbench,ding2025iworld}, but they do not provide an automated attack-search framework for closed-loop RL world-model agents. 
Meanwhile, static adversarial evaluation protocols such as AutoAttack \citep{croce2020autoattack,liang2022parallel,liang2025hard} reduce manual tuning for supervised vision models, but do not address the high-dimensional attack-configuration space and rollout cost of world-model evaluation. 
Recent autoresearch-style systems such as Claudini \citep{panfilov2026claudini,wang2025text} show that feedback-driven agents can discover strong adversarial attacks for LLMs, but their setting differs from closed-loop, dynamics-driven world models. 
WMAttack bridges this gap by treating adversarial evaluation as automated finite-budget attack search, using cross-task retrieval and feedback-driven refinement to identify high-impact configurations for world models.

\section{Preliminaries}

\subsection{Victim and Threat Model}

We consider adversarial evaluation of a trained world-model agent under closed-loop interaction with an environment. 
The victim agent is fixed during evaluation and is denoted by
\begin{equation}
    M_\theta = (\phi_\theta, f_\theta, \pi_\theta),
\end{equation}
where $\phi_\theta$ is the encoder or representation module, $f_\theta$ is the latent dynamics model, and $\pi_\theta$ is the policy. 
At each time step, the agent maps observations into latent states and selects actions from these internal representations. 
This abstraction covers recurrent state-space world models~\citep{hafner2021dreamerv2,hafner2023dreamerv3}, Transformer-based world models~\citep{micheli2023iris}, and latent dynamics models~\citep{hansen2024tdmpc2}.

The attack is applied only at evaluation time. 
The attacker does not modify the environment dynamics, reward function, or world-model parameters. 
Instead, adversarial influence enters the closed-loop trajectory only through perturbed observations received by the agent. 
Thus, a perturbation can affect long-horizon behavior only by propagating through the encoder, latent dynamics, and policy. 
Additional details on the clean and attacked closed-loop processes are provided in Appendix~\ref{app:victim-threat-model}.

\subsection{Adversary Capabilities}

For an evaluation task $\tau$, let $\mathcal{C}_\tau$ denote the attack configuration space. 
An attack configuration $c \in \mathcal{C}_\tau$ specifies both the attack family and its evaluation hyperparameters:
\begin{equation}
    c =
    (
    \text{attack type},
    \epsilon,
    \text{steps},
    \text{restarts},
    \rho,
    \text{seed},
    \text{allocation rule}
    ),
\end{equation}
where $\epsilon$ is the perturbation budget, $\text{steps}$ and $\text{restarts}$ control attack optimization, $\rho$ controls the adaptive step-size schedule when applicable, and the allocation rule specifies how attack resources are assigned during evaluation. 
In our implementation, the attack family is selected from APGD-CE and APGD-DLR~\citep{croce2020autoattack}, FAB~\citep{croce2020fab}, Square Attack~\citep{andriushchenko2020square}, and Physics-Conditioned World-Model Attack (PhysCond-WMA)~\citep{guo2026physcondwma}.
Detailed hyperparameter ranges are given in Appendix~\ref{app:exp-details}.

We consider bounded observation-space attacks on the image keys used by the world-model encoder. 
Let $\mathcal{K}_{\mathrm{img}}$ denote these keys. 
For each $k \in \mathcal{K}_{\mathrm{img}}$, the attacker replaces the clean observation with
\begin{equation}
    \tilde{o}_{t,k}
    =
    \Pi_{[-0.5,0.5]}
    \left(
    o_{t,k}+\delta_{t,k}
    \right),
    \qquad
    \|\delta_{t,k}\|_\infty \leq \epsilon/255,
\end{equation}
where $\Pi_{[-0.5,0.5]}$ projects the perturbed observation back to the normalized input range. 
Different attack families may use different optimization procedures or access assumptions, but WMAttack evaluates all of them under the same configuration space and utility function.

\subsection{Attack Goal}
\label{sec:attack-goal}
The goal of WMAttack is not to certify robustness or guarantee a global optimum, but to efficiently discover high-impact attacks under a finite evaluation budget. 
Let $J_{\mathrm{clean}}(\tau)$ denote the expected episodic return under clean observations, and let $J_{\mathrm{adv}}(\tau,c)$ denote the attacked return under configuration $c$. 
WMAttack scores each configuration by a scalarized utility:
\begin{equation}
    U_\tau(c)
    =
    D_\tau(c)
    +
    w_f F_\tau(c)
    -
    w_r \log(1+T_\tau(c))
    -
    w_v V_\tau(c),
\end{equation}
where $D_\tau(c)$ is normalized reward degradation, $F_\tau(c)$ measures action instability, $T_\tau(c)$ is evaluation time, $V_\tau(c)$ measures rollout variability, and $w_f,w_r,w_v$ balance these terms. 
The exact definitions of the utility components are provided in Appendix~\ref{app:utility-details}.

Under a finite budget $B$, adversarial evaluation is formulated as attack-configuration search:
\begin{equation}
    \max_{c_1,\ldots,c_B \in \mathcal{C}_\tau}
    \mathbb{E}
    \left[
    \max_{1 \leq i \leq B}
    U_\tau(c_i)
    \right].
\end{equation}
This formulation reflects the practical cost of world-model evaluation, where each candidate configuration may require closed-loop rollouts and repeated latent-dynamics computation.

\section{Method}
\subsection{Problem Decomposition}

WMAttack takes the finite-budget attack-search objective in Section~\ref{sec:attack-goal} as its starting point and learns how to allocate the limited evaluation budget across $\mathcal{C}_\tau$.
Instead of sampling configurations independently from a fixed prior, WMAttack maintains an adaptive proposal distribution $q_t(c\mid\tau)$ over attack configurations. 
This distribution represents the current belief about which configurations are likely to produce high utility on task $\tau$.

At the beginning of search, the main challenge is cold-start initialization: for a new task-agent pair, the evaluator does not know which attack family or hyperparameter region is likely to be effective. 
WMAttack addresses this by using \textbf{Representation-Guided Attack Retrieval} (RGAR) to initialize $q_0(c\mid\tau)$ from historical configurations that were effective on representation-similar tasks. 
After each evaluation round, the challenge becomes adaptation: the method must use trial outcomes to refine the search direction. 
WMAttack addresses this with \textbf{Self-Correcting Attack Search} (SCAS), which updates $q_t(c\mid\tau)$ using feedback from reward degradation, action instability, evaluation cost, and rollout variability.
Thus, the finite-budget search problem is decomposed into two coupled steps:
\begin{equation}
    \text{RGAR: } \mathcal{M}, \psi_\tau \mapsto q_0(c\mid\tau),
    \qquad
    \text{SCAS: } q_t(c\mid\tau), \mathcal{H}_t \mapsto q_{t+1}(c\mid\tau),
\end{equation}
where $\mathcal{M}$ is the attack memory, $\psi_\tau$ is the latent behavioral summary of the new task-agent pair, and $\mathcal{H}_t$ is the accumulated evaluation history. 
The next two sections describe these two steps in detail.

\subsection{Representation-Guided Attack Retrieval}
\begin{figure*}[t]
    \centering
    \includegraphics[width=\textwidth]{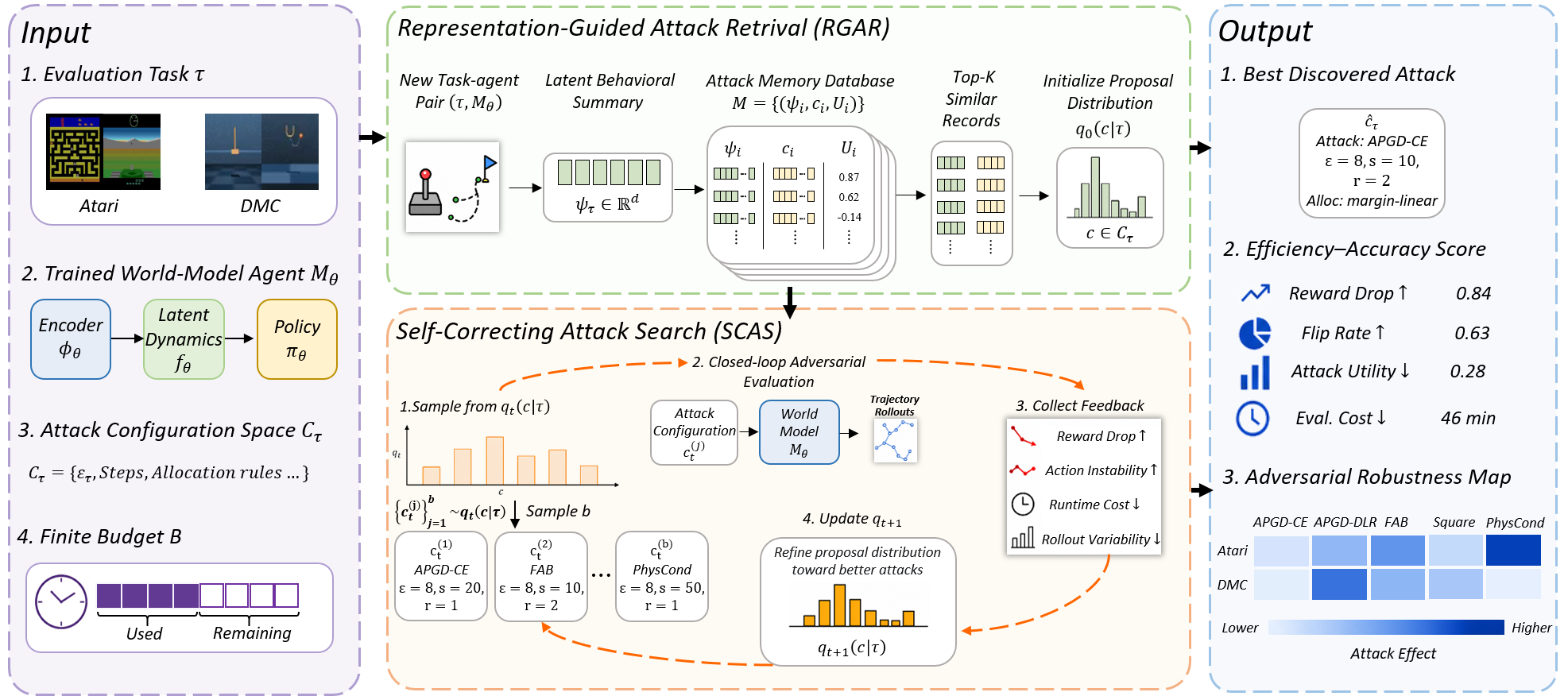}
    \caption{
    Overview of WMAttack. RGAR initializes the attack proposal distribution from representation-similar historical tasks, and SCAS refines it using closed-loop rollout feedback.
    }
    \label{fig:wmbench-overview}
\end{figure*}
A central challenge in attacking world-model agents is that attack effectiveness varies substantially across tasks, models, and environments. 
However, agents with similar latent dynamics or decision behaviors may share effective attack configurations~\citep{behzadan2017vulnerability, zhang2021learning,wang2023diversifying}. 
Tasks with similar temporal dependencies, reward sensitivities, or latent transition patterns may respond similarly to perturbation budgets, optimization steps, or allocation rules. 
RGAR exploits this observation by retrieving prior adversarial experience through latent behavioral representations~\citep{liu2023improving,liu2025bridging}.

For each task-agent pair, WMAttack constructs a latent behavioral summary
\begin{equation}
    \psi_\tau
    =
    g
    \left(
    \{z_t, \hat{z}_{t+k}, u_t, r_t\}_{t=1}^{H}
    \right),
\end{equation}
where $z_t$ denotes the agent's latent state, $\hat{z}_{t+k}$ denotes predicted future latent states, $u_t$ is the selected action, and $r_t$ is the observed reward. 
The function $g(\cdot)$ aggregates trajectory-level information into a compact representation that characterizes the behavior of the agent under clean evaluation.

Given a new task summary $\psi_\tau$, RGAR retrieves the top-$K$ most relevant historical records:
\begin{equation}
    \mathcal{N}_K(\tau)
    =
    \mathrm{TopK}_{i}
    \; \mathrm{sim}(\psi_\tau,\psi_i),
\end{equation}
where $\mathrm{sim}(\cdot,\cdot)$ measures similarity in the latent representation space. 
The retrieved configurations are then converted into a warm-start proposal distribution:
\begin{equation}
    q_0(c \mid \tau)
    =
    (1-\lambda)q_{\mathrm{base}}(c)
    +
    \lambda
    \sum_{i \in \mathcal{N}_K(\tau)}
    \alpha_i \delta(c=c_i),
\end{equation}
where $q_{\mathrm{base}}$ is a generic prior over attack configurations, $\delta(\cdot)$ is a point mass on a retrieved configuration, and $\lambda \in [0,1]$ controls the strength of retrieval. 
The weights $\alpha_i$ are assigned according to retrieval similarity and historical attack utility.
This retrieval stage reduces cold-start search. 
Rather than initializing from uninformed random configurations, WMAttack starts from configurations that were effective for representation-similar task-agent pairs, improving the initial attack proposal distribution before task-specific feedback is collected.

\subsection{Self-Correcting Attack Search}

RGAR provides an informed warm start, but the retrieved configurations may not be optimal for a new task. 
SCAS therefore refines the search process using feedback from evaluated attack trials. 
We view the ranked candidate generator as an implicit proposal distribution $q_t(c \mid \tau)$ over the attack configuration space.
At round $t$, WMAttack proposes a batch of candidate configurations and evaluates them through closed-loop rollouts. 
Each candidate is scored by the scalarized utility $U_\tau(c)$ defined in Section~\ref{sec:attack-goal}. 
In implementation, WMAttack uses a scout-confirm protocol: inexpensive scout rollouts estimate candidate utility, and the most promising candidates are re-evaluated with more episodes for confirmation. 
The search history is updated as
\begin{equation}
    \mathcal{H}_t
    =
    \mathcal{H}_{t-1}
    \cup
    \{(c_t^{(j)}, U_\tau(c_t^{(j)}), A_t^{(j)})\}_{j=1}^{b},
\end{equation}
where $A_t^{(j)}$ denotes the feedback signal for candidate $c_t^{(j)}$, including observed failure patterns, parameter recommendations, and search-direction suggestions.

SCAS uses this feedback to generate and rank new candidates. 
High-utility trials serve as anchors for local refinement, while feedback signals adjust the search direction by changing $\epsilon$, optimization steps, restarts, or the allocation rule. 
Abstractly, the proposal update is written as
\begin{equation}
    q_{t+1}(c \mid \tau)
    =
    (1-\alpha_t) q_t(c \mid \tau)
    +
    \alpha_t \hat{q}_t(c \mid \mathcal{H}_t),
\end{equation}
where $\alpha_t$ is the update rate and $\hat{q}_t$ is the feedback-induced proposal distribution estimated from the accumulated search history. 
In implementation, this update is realized through candidate generation, local neighborhood expansion, feedback-guided variants, and ranked selection rather than explicit probabilistic sampling. 
This self-correcting loop concentrates the evaluation budget on regions of the attack space that are more likely to expose meaningful world-model failures.

\subsection{Theoretical Analysis}

We provide a concise explanation for why RGAR and SCAS can improve finite-budget attack search. 
The full derivation, including the construction of the utility-biased reference distribution, noisy correction, comparison with a Claudini-style baseline, and finite-episode utility estimation, is deferred to Appendix~\ref{app:theory}.

Let $\mathcal{G}_{\eta}$ be the set of $\eta$-effective attack configurations:
\begin{equation}
    \mathcal{G}_{\eta}
    =
    \{c \in \mathcal{C}_\tau :
    U_\tau(c) \geq U_\tau^\star - \eta\},
    \qquad
    U_\tau^\star = \max_{c \in \mathcal{C}_\tau} U_\tau(c).
\end{equation}
At round $t$, let $q_t$ be the implicit proposal distribution induced by retrieval context, candidate ranking, and search feedback, and define $p_t=q_t(\mathcal{G}_\eta)$ as the probability of sampling an effective configuration.

We model self-correcting proposal refinement as a distribution-level update
\begin{equation}
    C_\gamma(q)
    =
    (q+\gamma q^\star)/(1+\gamma),
\end{equation}
where $\gamma \geq 0$ is the correction strength and $q^\star$ is a utility-biased reference distribution, instantiated in Appendix~\ref{app:theory} as $q_{\tau,\beta}^\star(c)\propto \exp(\beta U_\tau(c))$. 
For the effective set, this update gives
\begin{equation}
    C_\gamma(q)(\mathcal{G}_\eta)-q(\mathcal{G}_\eta)
    =
    \gamma\big(q^\star(\mathcal{G}_\eta)-q(\mathcal{G}_\eta)\big)/(1+\gamma).
\end{equation}
Thus, self-correction increases the probability of sampling an effective attack whenever the reference distribution is more concentrated on $\mathcal{G}_\eta$ than the current proposal distribution. 
As a clean special case, if $q^\star(\mathcal{G}_\eta)=1$, the probability mass outside $\mathcal{G}_\eta$ contracts by a factor of $1/(1+\gamma)$.

This mass improvement translates into finite-budget search efficiency. 
If a round samples $b$ candidates from $q_t$, the probability that the batch contains at least one effective configuration is
\begin{equation}
    h_t = 1-(1-p_t)^b.
\end{equation}
Since $h_t$ is monotone increasing in $p_t$, any update that increases $q_t(\mathcal{G}_\eta)$ improves the probability of discovering an effective attack in the next batch. 
If $p_t \geq p>0$ across rounds, the expected hitting time satisfies
\begin{equation}
    \mathbb{E}[T]
    \leq
    1/\big(1-(1-p)^b\big).
\end{equation}

Therefore, the theory does not claim global attack optimization. 
It shows that RGAR and SCAS improve finite-budget attack search whenever they shift sufficient proposal mass toward high-utility attack configurations, with noisy and finite-sample cases treated in Appendix~\ref{app:theory}.

\section{Experiments}
\subsection{Experimental Setup}

\textbf{Datasets and victim models.}
We evaluate \textbf{WMAttack} on Atari~\citep{bellemare2013arcade} and DeepMind Control Suite (DMC)~\citep{tassa2018deepmind}. 
The main experiments use DreamerV3~\citep{hafner2023dreamerv3} as the primary victim model, covering 26 Atari tasks and 20 DMC tasks. 
To examine cross-model applicability, we further report additional results on DreamerV2~\citep{hafner2021dreamerv2}, TD-MPC2~\citep{hansen2024tdmpc2}, and IRIS~\citep{micheli2023iris}. 
All victim agents are trained before evaluation and remain fixed during adversarial testing.
\\\textbf{Attack search space.}
WMAttack searches over multiple attack families: APGD-CE and APGD-DLR~\citep{croce2020autoattack}, FAB~\citep{croce2020fab}, Square Attack~\citep{andriushchenko2020square}, and PhysCond-WMA~\citep{guo2026physcondwma}. 
For each task and attack family, the configuration space includes the perturbation budget $\epsilon$, optimization steps, restarts, $\rho$, random seed, and allocation rule. 
Each candidate configuration is evaluated through closed-loop rollouts and scored using the scalarized utility defined in Section~\ref{sec:attack-goal}. 
All compared methods use the same victim checkpoints, search space, rollout protocol, and evaluation budget within each setting.
\\\textbf{Baselines and ablations.}
We compare WMAttack against two baselines. 
\textbf{Random} samples attack configurations without retrieval or feedback-conditioned refinement. 
\textbf{Claudini} is a Claudini-style autoresearch baseline adapted from Claudini~\citep{panfilov2026claudini}; it uses iterative feedback to propose attack configurations but does not use RGAR or cross-task latent retrieval. 
We also conduct ablations to isolate RGAR and evaluate different LLM backbones for proposal generation, including Qwen~\citep{qwen2024qwen25} and LLaMA~\citep{dubey2024llama}.
\\\textbf{Metrics.}
We report both attack strength and search efficiency. 
\textbf{Drop} is normalized reward degradation, \textbf{Flip} is the action flip rate, and \textbf{Utility} is the scalarized attack objective. 
\textbf{Time} reports wall-clock evaluation time, while threshold-based efficiency measures how often a method reaches a target fraction of its final best utility. 
Detailed task lists, hyperparameter ranges, rollout budgets, and implementation details are provided in Appendix~\ref{app:exp-details}.
\begin{table}[t]
\centering
\caption{Main adversarial evaluation results. }
\label{tab:main-results}
\small
\begin{tabular}{lcccccccc}
\toprule
Victim & Suite & Tasks & Attacks & Method & Drop $\uparrow$ & Flip $\uparrow$ & Utility $\uparrow$ & Time (min) \\
\midrule
DreamerV3 & Atari & 26 & 4 & Claudini & 0.497 & 0.400 & -0.132 & 77.2 \\
DreamerV3 & Atari & 26 & 4 & Random & 0.446 & 0.377 & -0.163 & 73.4 \\
DreamerV3 & Atari & 26 & 4 & WMAttack & \textbf{1.034} & \textbf{0.660} & \textbf{0.430} & 106.2 \\
\midrule
DreamerV3 & DMC & 20 & 5 & Claudini & 0.319 & 0.705 & -0.137 & 55.4 \\
DreamerV3 & DMC & 20 & 5 & Random & 0.352 & 0.721 & -0.123 & 59.0 \\
DreamerV3 & DMC & 20 & 5 & WMAttack & \textbf{0.682} & \textbf{0.800} & \textbf{0.174} & 120.2 \\
\midrule
DreamerV2 & Atari & 10 & 4 & WMAttack & 0.539 & 0.554 & -0.021 & 242.9 \\
DreamerV2 & DMC & 10 & 5 & WMAttack & 0.397 & 0.762 & 0.046 & 253.5 \\
TD-MPC2 & DMC & 10 & 5 & WMAttack & 0.621 & 0.786 & 0.152 & 68.1 \\
IRIS & Atari & 10 & 4 & WMAttack & 0.550 & 0.474 & -0.214 & 830.1 \\
\bottomrule
\end{tabular}
\end{table}
\begin{figure}[t]
    \centering
    \begin{minipage}[t]{0.49\linewidth}
        \centering
        \includegraphics[width=\linewidth]{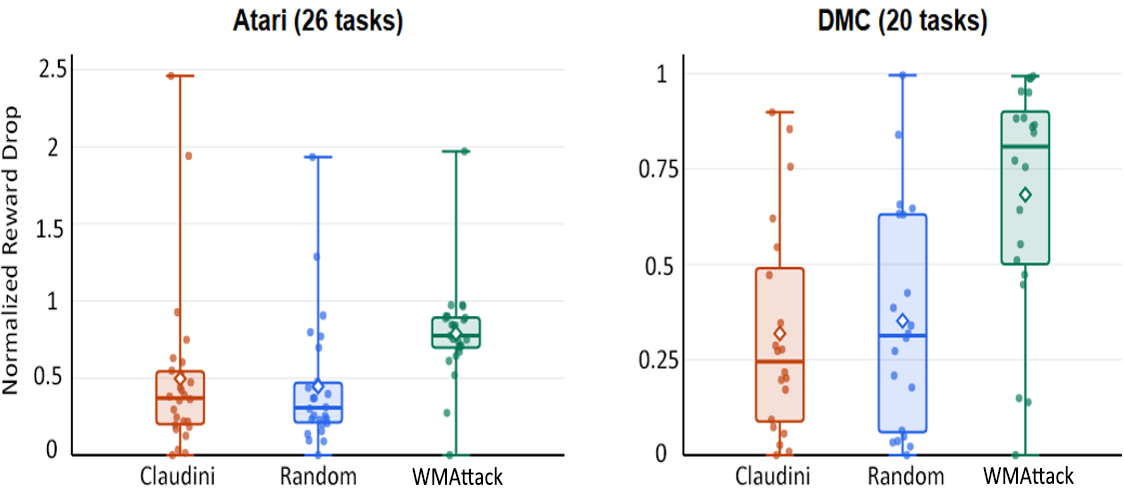}
        \centerline{\small (a) Reward-drop distribution}
    \end{minipage}
    \hfill
    \begin{minipage}[t]{0.49\linewidth}
        \centering
        \includegraphics[width=\linewidth]{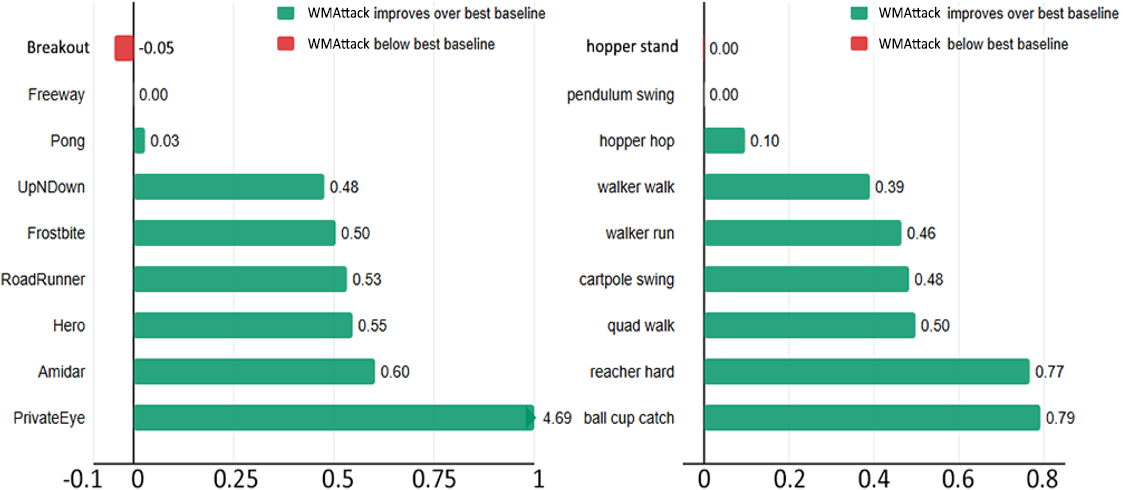}
        \centerline{\small (b) Per-task improvement}
    \end{minipage}
    \caption{Task-level reward-drop distribution and per-task improvement on DreamerV3 Atari and DMC.
    WMAttack shifts the reward-drop distribution upward and improves over the stronger baseline on most displayed tasks.}
    \label{fig:task-level-results}
\end{figure}

\subsection{Main Results}
\begin{table}[t]
\centering
\caption{
Threshold-based search efficiency.
Hit Rate reports the fraction of task--attack pairs that reach $90\%$ of their final best utility.
}
\label{tab:efficiency-threshold}
\small
\begin{tabular}{lllrccc}
\toprule
Victim & Suite & Method & Pairs & Hit Rate $\uparrow$ & Trials $\downarrow$ & Time (s) \\
\midrule
DreamerV3 & Atari & Claudini & 104 & 0.538 & 3.36 & 349 \\
DreamerV3 & Atari & Random & 104 & 0.529 & 3.67 & 394 \\
DreamerV3 & Atari & WMAttack & 104 & \textbf{0.875} & \textbf{3.45} & 491 \\
\midrule
DreamerV3 & DMC & Claudini & 100 & 0.170 & 3.53 & 254 \\
DreamerV3 & DMC & Random & 100 & 0.180 & \textbf{3.44} & 325 \\
DreamerV3 & DMC & WMAttack & 100 & \textbf{0.490} & 3.71 & 823 \\
\midrule
DreamerV2 & Atari & WMAttack & 40 & 0.545 & 2.47 & 111 \\
DreamerV2 & DMC & WMAttack & 50 & 0.213 & 2.69 & 232 \\
TD-MPC2 & DMC & WMAttack & 50 & 0.550 & 2.27 & 339 \\
IRIS & Atari & WMAttack & 40 & 0.000 & -- & -- \\
\bottomrule
\end{tabular}
\end{table}
\begin{figure}[t]
    \centering
    \begin{minipage}[t]{0.49\linewidth}
        \centering
        \includegraphics[width=\linewidth]{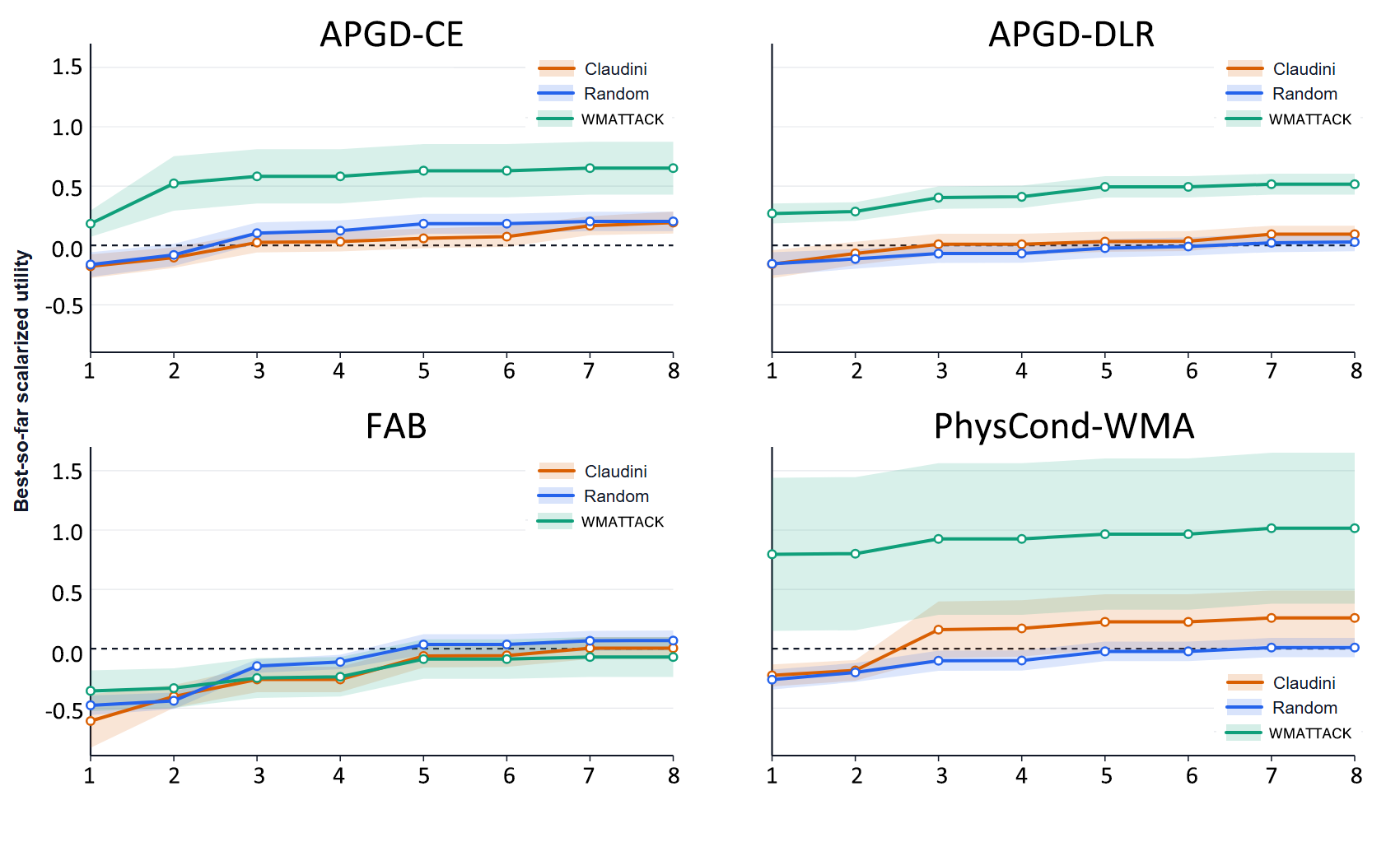}
        \vspace{1mm}
        \centerline{\small (a) Atari}
    \end{minipage}
    \hfill
    \begin{minipage}[t]{0.49\linewidth}
        \centering
        \includegraphics[width=\linewidth]{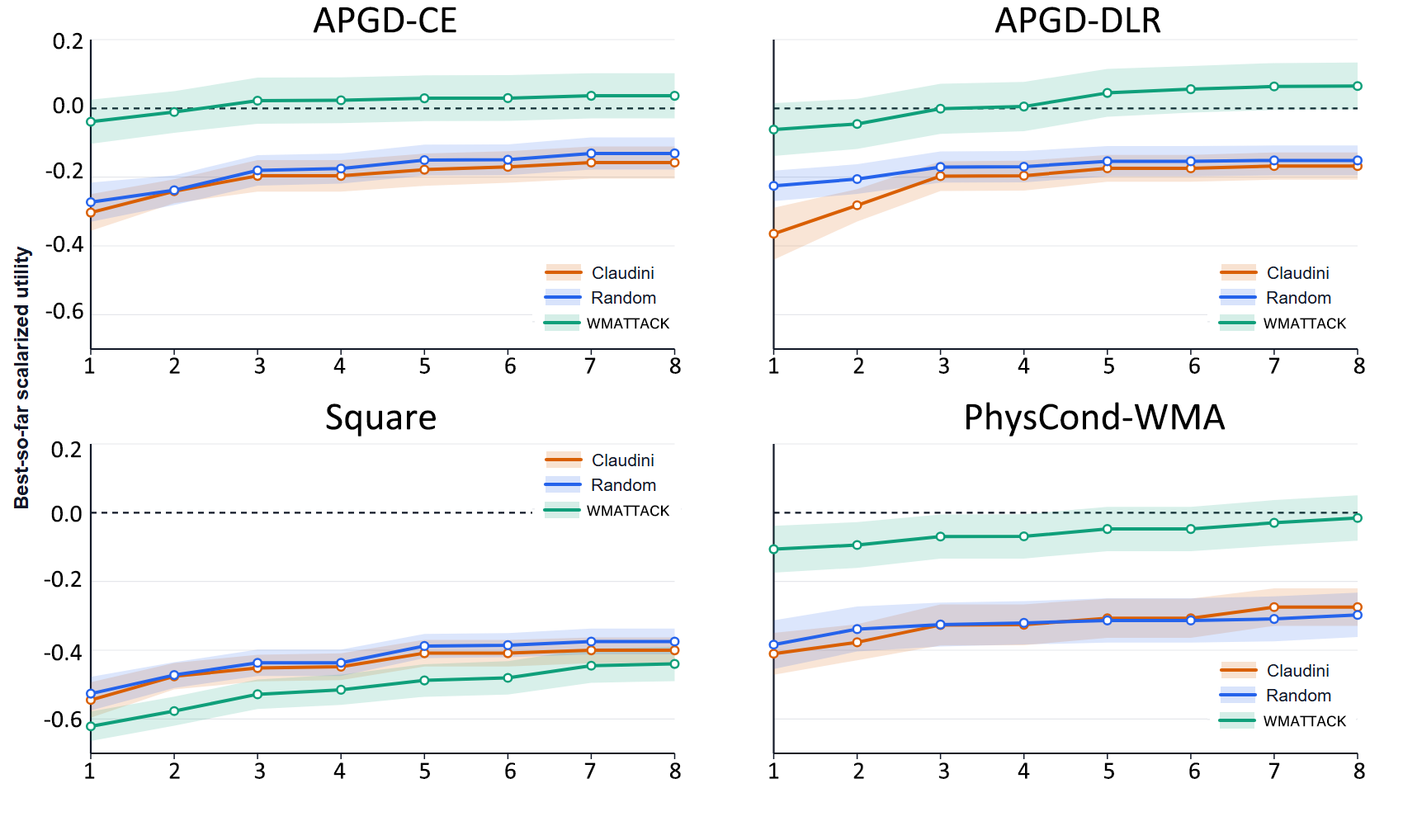}
        \vspace{1mm}
        \centerline{\small (b) DMC}
    \end{minipage}
    \vspace{1mm}
    \caption{
    Attack-specific search efficiency on DreamerV3 Atari and DMC.
    WMAttack generally reaches higher best-so-far utility earlier than Claudini and Random across most attack families.
    }
    \label{fig:efficiency-curves}
\end{figure}

Table~\ref{tab:main-results} summarizes the primary adversarial evaluation results across multiple suites and model architectures. Following \citet{panfilov2026claudini}, we compare WMAttack against two baselines: \textit{Random}, which performs search without retrieval or feedback-conditioned refinement, and an adapted \textit{Claudini}-style autoresearch baseline.
For DreamerV3 Atari, the main results use four attack families because Square Attack is substantially more expensive in the Atari closed-loop setting and was therefore excluded from the primary Atari runs; DMC includes all five attack families.
\paragraph{Superior Adversarial Potency on DreamerV3.} WMAttack significantly outperforms both baselines on the DreamerV3 agent. On Atari, WMAttack achieves an average normalized reward drop of $1.034$, surpassing Random and Claudini by $131.8\%$ and $108.0\%$, respectively. Notably, WMAttack is the only method to yield a positive scalarized utility $0.430$, indicating that the strength of the discovered attacks justifies the search cost. Similar trends are observed on DMC, where WMAttack reaches a reward drop of $0.682$ and the highest action flip rate $0.800$. These results validate that the synergistic combination of RGAR and SCAS identifies substantially more potent adversarial configurations than either random exploration or feedback-only search.
\paragraph{Cross-model generalization.}
We further evaluate WMAttack on DreamerV2, TD-MPC2, and IRIS to test whether the attack-search framework extends beyond the primary DreamerV3 setting. 
WMAttack exposes clear vulnerabilities across these model families, achieving reward drops of $0.539$ on DreamerV2 Atari, $0.397$ on DreamerV2 DMC, $0.621$ on TD-MPC2 DMC, and $0.550$ on IRIS Atari. 
The scalarized utility differs across victims because it also accounts for evaluation cost; for example, IRIS obtains a high reward drop but negative utility due to its much larger rollout runtime. 
These results suggest that WMAttack can be applied to different world-model architectures, while also revealing architecture-dependent strength--cost trade-offs.

\paragraph{Consistency and attack-family selection.}
Figure~\ref{fig:task-level-results} provides a task-level view of the DreamerV3 results. 
Panel (a) shows that WMAttack shifts the reward-drop distribution upward relative to both baselines, indicating stronger attacks across the task distribution. 
Panel (b) further shows that WMAttack improves over the stronger baseline on most tasks, achieving the highest reward drop on 24 out of 26 Atari tasks and 18 out of 20 DMC tasks. 
These results indicate that the gains are consistent across environments rather than driven by a small number of outliers. 
The strongest attack family also varies across suites: APGD-CE is selected most often on Atari, whereas FAB dominates on DMC. 
This heterogeneity motivates searching over multiple attack families instead of relying on a single fixed attack. 
Detailed attack-family selection statistics are provided in Appendix~\ref{app:attack-family-selection}.

\subsection{Efficiency--Accuracy Trade-off}
\begin{table}[t]
\centering
\caption{
Ablation of RGAR and SCAS on DreamerV3 Atari.
Each cell reports Utility / Drop / Flip.
}
\label{tab:rgar-ablation}
\small
\begin{tabular}{lcc}
\toprule
Variant & First Utility / Drop / Flip $\uparrow$ & Final Utility / Drop / Flip $\uparrow$ \\
\midrule
SCAS w/o RGAR 
& 0.048 / 0.445 / 0.394 
& -0.325 / 0.349 / 0.396 \\
RGAR + SCAS 
& \textbf{0.397 / 0.770 / 0.573} 
& \textbf{0.020 / 0.665 / 0.568} \\
\midrule
RGAR only, w/o SCAS  
& 0.245 / 0.744 / 0.580
& -0.046 / 0.645 / 0.572 \\
w/o RGAR and SCAS 
& 0.214 / 0.687 / 0.604
& -0.048 / 0.615 / 0.602 \\
\bottomrule
\end{tabular}
\end{table}
\begin{table}[t]
\centering
\caption{
Sensitivity to the LLM backbone on DreamerV3 Atari.}
\label{tab:llm-backbone}
\small
\begin{tabular}{llccccc}
\toprule
Model & Size & Drop $\uparrow$ & Flip $\uparrow$ & Utility $\uparrow$ & Trials $\downarrow$ & Time (s) \\
\midrule
 & 1B & 0.715 & 0.637 & 0.114 & 12.18 & 5528 \\
LLaMA & 3B & 0.812 & 0.668 & 0.210 & 12.67 & 6803 \\
 & 8B & 0.769 & 0.629 & 0.171 & 12.46 & 5215 \\
\midrule
 & 1.5B & 0.754 & 0.640 & 0.168 & \textbf{9.76} & \textbf{5164} \\
Qwen & 3B & \textbf{1.034} & 0.660 & \textbf{0.430} & 10.66 & 6338 \\
 & 7B & 0.916 & \textbf{0.675} & 0.314 & 10.45 & 8252 \\
\bottomrule
\end{tabular}
\end{table}
We evaluate whether WMAttack can find strong attacks efficiently under a limited search budget. 
Table~\ref{tab:efficiency-threshold} reports the fraction of task--attack pairs that reach $90\%$ of their final best utility, while detailed attack-specific efficiency curves are provided in Appendix~\ref{app:efficiency-details}.
\\\textbf{WMAttack enters high-utility regions more reliably.}
On DreamerV3 Atari, WMAttack reaches the threshold on $87.5\%$ of task--attack pairs, compared with $52.9\%$ for Random and $53.8\%$ for Claudini. 
On DMC, the hit rate improves from about $18\%$ for the baselines to $49.0\%$. 
Although WMAttack can require more wall-clock time, especially on DMC, it substantially increases the probability of finding useful attacks under the same finite search protocol.
\\\textbf{The efficiency gains transfer beyond DreamerV3.}
WMAttack also reaches nontrivial hit rates on DreamerV2 Atari ($54.5\%$), DreamerV2 DMC ($21.3\%$), and TD-MPC2 DMC ($55.0\%$). 
For IRIS, the hit rate is zero under this threshold because its final utilities remain negative after accounting for the high rollout cost, even though Table~\ref{tab:main-results} shows nontrivial reward degradation.

\subsection{Ablation and Design Analysis}
\textbf{RGAR and SCAS are complementary.}
Table~\ref{tab:rgar-ablation} shows that adding RGAR to SCAS improves first-round Utility / Drop / Flip from $0.048/0.445/0.394$ to $0.397/0.770/0.573$, confirming that representation-guided retrieval provides a stronger warm start. 
After search, the full \emph{RGAR + SCAS} variant achieves the best final Utility / Drop / Flip $0.020/0.665/0.568$, while \emph{RGAR only} and \emph{w/o RGAR and SCAS} still have negative final utility. 
This suggests that retrieval improves initialization, but feedback-driven refinement is needed to convert a good warm start into higher final attack utility. 
Additional attack-family and pair-level analyses are provided in Appendix~\ref{app:ablation-details}.
\\\textbf{Larger LLM backbones are not uniformly better.}
Table~\ref{tab:llm-backbone} shows that attack-search performance does not scale monotonically with model size. 
Qwen2.5-3B obtains the best reward drop ($1.034$) and utility ($0.430$), Qwen2.5-7B obtains the highest flip rate ($0.675$) but incurs the largest runtime, and Qwen2.5-1.5B is the most efficient but weaker in final attack strength. 
These results suggest that the LLM backbone acts as a proposal generator, and its effectiveness depends on the strength--cost trade-off rather than parameter count alone.

\section{Conclusion}

We introduced \textbf{WMAttack}, an automated attack-search framework for adversarial evaluation of world-model agents. 
By combining representation-guided retrieval with self-correcting proposal refinement, WMAttack searches high-impact attack configurations under finite evaluation budgets. 
Experiments on Atari and DMC show stronger attacks than random search and a Claudini-style baseline, while ablations confirm the benefits of RGAR and feedback-conditioned search.

\bibliographystyle{plainnat}
\bibliography{neurips_2026}


\appendix
\clearpage
\section{Victim and Threat Model Details}
\label{app:victim-threat-model}

This appendix details the closed-loop evaluation process under clean and attacked observations. 
The victim is a fixed world-model agent
\begin{equation}
    M_\theta = (\phi_\theta, f_\theta, \pi_\theta),
\end{equation}
where $\phi_\theta$ is the encoder or representation module, $f_\theta$ is the latent dynamics model, and $\pi_\theta$ is the policy.

Under clean evaluation, the agent interacts with the environment as
\begin{equation}
    z_t = \phi_\theta(o_{\leq t}, u_{<t}),
    \qquad
    u_t \sim \pi_\theta(\cdot \mid z_t),
    \qquad
    o_{t+1}, r_t \sim \mathcal{E}(\cdot \mid o_t, u_t),
\end{equation}
where $\mathcal{E}$ denotes the environment dynamics and reward process. 
Under adversarial evaluation, the environment dynamics remain unchanged, but the agent receives perturbed observations:
\begin{equation}
    \tilde{z}_t = \phi_\theta(\tilde{o}_{\leq t}, \tilde{u}_{<t}),
    \qquad
    \tilde{u}_t \sim \pi_\theta(\cdot \mid \tilde{z}_t),
    \qquad
    o_{t+1}, r_t^{c} \sim \mathcal{E}(\cdot \mid o_t, \tilde{u}_t).
\end{equation}
Thus, attacks affect returns only indirectly by changing the observations used to construct latent states and actions.

In WMAttack, attacks are applied to image observation keys used by the world-model encoder. 
Let $\mathcal{K}_{\mathrm{img}}$ denote this set. 
For each $k \in \mathcal{K}_{\mathrm{img}}$, the attacked observation is
\begin{equation}
    \tilde{o}_{t,k}
    =
    \Pi_{[-0.5,0.5]}
    \left(
    o_{t,k} + \delta_{t,k}
    \right),
    \qquad
    \|\delta_{t,k}\|_\infty \leq \epsilon/255.
\end{equation}
The projection keeps inputs within the normalized observation range, and $\epsilon$ is specified in pixel scale. 
Non-image keys are left unchanged:
\begin{equation}
    \tilde{o}_{t,k}=o_{t,k},
    \qquad
    k \notin \mathcal{K}_{\mathrm{img}}.
\end{equation}

The attacker may choose the attack configuration $c\in\mathcal{C}_\tau$, including the attack family and hyperparameters, but cannot modify the environment dynamics, reward function, policy parameters, world-model parameters, checkpoint, or training data.

\clearpage
\section{Attack Configuration Space}
\label{app:exp-details}

This appendix provides additional details on the attack configuration space used by WMAttack. 
For each task and attack family, WMAttack searches over a discrete set of attack configurations. 
Each configuration contains the attack type, perturbation budget $\epsilon$, number of optimization steps, number of restarts, $\rho$, random seed, and allocation rule. 
The perturbation budget $\epsilon$ is specified in pixel scale and converted to $\epsilon/255$ in the normalized observation space.

WMAttack evaluates five attack families:
\begin{itemize}
    \item \textbf{APGD-CE}: an adaptive projected-gradient attack using cross-entropy loss.
    \item \textbf{APGD-DLR}: an adaptive projected-gradient attack using a difference-of-logits-ratio style objective.
    \item \textbf{FAB}: a boundary-oriented attack adapted to the world-model policy output.
    \item \textbf{Square Attack}: a black-box patch-based attack that searches over square perturbation regions.
    \item \textbf{PhysCond-WMA}: a physics-conditioned world-model attack designed to exploit world-model-specific latent and policy sensitivities.
\end{itemize}

The search spaces used in our implementation are summarized in Table~\ref{tab:attack-search-space}. 
Exact values may be adjusted for different hardware or evaluation budgets, but all compared methods use the same search space.

\begin{table}[h]
\centering
\caption{Attack search spaces used by WMAttack.}
\label{tab:attack-search-space}
\begin{tabular}{llll}
\toprule
Attack & $\epsilon$ candidates & Step candidates & Allocation modes \\
\midrule
APGD-CE & $2$--$20$ & $4$--$24$ & fixed, margin-linear \\
APGD-DLR & $2$--$20$ & $4$--$24$ & fixed, margin-linear \\
FAB & $2$--$20$ & $6$--$32$ & fixed, margin-linear \\
PhysCond-WMA & $2$--$20$ & $6$--$32$ & fixed, margin-linear \\
Square & $2$--$16$ & $20$--$160$ & fixed, margin-linear \\
\bottomrule
\end{tabular}
\end{table}

For the fixed allocation mode, WMAttack applies the same attack budget at each evaluated decision point. 
For margin-linear allocation, WMAttack adaptively adjusts the effective number of attack steps and perturbation budget according to the policy margin. 
This allows the evaluator to allocate more attack effort to states where the policy is more confident and less effort to states where the action is already unstable.

The remaining configuration fields, including restarts, $\rho$, and random seeds, follow the evaluator configuration used for each run. 
They are kept consistent across WMAttack and all baselines to ensure that performance differences come from the search strategy rather than from different attack spaces.

\clearpage
\section{Utility Components}
\label{app:utility-details}

This appendix provides the detailed definitions of the utility components used in WMAttack. 
For a task $\tau$, let $J_{\mathrm{clean}}(\tau)$ denote the expected episodic return under clean observations, and let $J_{\mathrm{adv}}(\tau,c)$ denote the expected attacked return under attack configuration $c$.

The normalized reward degradation is defined as
\begin{equation}
    D_\tau(c)
    =
    \frac{
    J_{\mathrm{clean}}(\tau) - J_{\mathrm{adv}}(\tau,c)
    }{
    |J_{\mathrm{clean}}(\tau)| + 1
    }.
\end{equation}
This term is larger when the attack causes a stronger drop in episodic return. 
The denominator normalizes the score across tasks with different reward scales and avoids division by zero when the clean return is close to zero.

The action-instability term $F_\tau(c)$ measures how often the adversarial observation changes the agent's selected action. 
For discrete action spaces, it is computed as the fraction of evaluated decision points where
\begin{equation}
    \arg\max_u \pi_\theta(u \mid z_t)
    \neq
    \arg\max_u \pi_\theta(u \mid \tilde{z}_t).
\end{equation}
For continuous action spaces, WMAttack uses a thresholded action shift:
\begin{equation}
    \mathbb{I}
    \left[
    \left\|
    u_t - \tilde{u}_t
    \right\|_1
    >
    \kappa
    \right],
\end{equation}
where $\kappa$ is a small threshold. 
These per-step measurements are averaged over rollout steps and episodes to obtain $F_\tau(c)$.

The runtime term $T_\tau(c)$ denotes the wall-clock time required to evaluate configuration $c$. 
It is penalized through $\log(1+T_\tau(c))$ so that expensive attacks are discouraged while avoiding an overly large penalty for moderate runtime differences.

The rollout-variability term is defined as
\begin{equation}
    V_\tau(c)
    =
    \frac{
    \mathrm{Std}[R_{\mathrm{adv}}(c)]
    }{
    |J_{\mathrm{clean}}(\tau)| + 1
    },
\end{equation}
where $R_{\mathrm{adv}}(c)$ denotes episodic returns from repeated attacked rollouts under configuration $c$. 
This term penalizes configurations whose outcomes are highly variable across repeated evaluations.

Combining these components, WMAttack uses the scalarized utility
\begin{equation}
    U_\tau(c)
    =
    D_\tau(c)
    +
    w_f F_\tau(c)
    -
    w_r \log(1 + T_\tau(c))
    -
    w_v V_\tau(c).
\end{equation}
In our implementation, we use $w_f=0.25$, $w_r=0.15$, and $w_v=0.05$. 
Higher utility corresponds to stronger reward degradation, higher action instability, lower evaluation time, and lower rollout variability.

\clearpage
\section{Theoretical Details}
\label{app:theory}

This appendix provides the derivations omitted from Section~4.4. 
We use the same notation as in the main text: $\mathcal{C}_\tau$ is the finite attack configuration space, $U_\tau(c)$ is the scalarized utility, and $\mathcal{G}_{\eta}$ is the set of $\eta$-effective configurations.

\subsection{Utility-Biased Reference Distribution}

Section~4.4 uses a reference distribution $q^\star$ that assigns larger probability to high-utility attacks. 
A concrete construction is the Gibbs distribution
\begin{equation}
    q_{\tau,\beta}^{\star}(c)
    =
    \frac{\exp(\beta U_\tau(c))}
    {\sum_{c' \in \mathcal{C}_\tau}\exp(\beta U_\tau(c'))},
    \qquad c\in\mathcal{C}_\tau,
\end{equation}
where $\beta\geq 0$ is an inverse-temperature parameter. 
When $\beta=0$, this distribution is uniform; as $\beta$ increases, it concentrates more mass on high-utility configurations. 
Let
\begin{equation}
    r
    =
    q_{\tau,\beta}^{\star}(\mathcal{G}_{\eta})
\end{equation}
denote the reference mass on the effective set.

For the self-correction operator
\begin{equation}
    C_\gamma(q)
    =
    (q+\gamma q^\star)/(1+\gamma),
\end{equation}
the effective-set mass is
\begin{equation}
    C_\gamma(q)(\mathcal{G}_{\eta})
    =
    \frac{q(\mathcal{G}_{\eta})+\gamma r}{1+\gamma}.
\end{equation}
Therefore,
\begin{equation}
    C_\gamma(q)(\mathcal{G}_{\eta})-q(\mathcal{G}_{\eta})
    =
    \frac{\gamma}{1+\gamma}
    \left(r-q(\mathcal{G}_{\eta})\right).
\end{equation}
Thus, correction improves the probability of sampling an effective attack whenever $r>q(\mathcal{G}_{\eta})$. 
The oracle case in the main text corresponds to $r=1$, which gives
\begin{equation}
    1-C_\gamma(q)(\mathcal{G}_{\eta})
    =
    \frac{1-q(\mathcal{G}_{\eta})}{1+\gamma}.
\end{equation}

\subsection{Noisy Correction}

The realized update need not exactly equal $C_\gamma(q_t)$. 
Suppose the next proposal distribution satisfies the sufficient noisy-correction condition
\begin{equation}
    q_{t+1}(\mathcal{G}_{\eta})
    \geq
    C_\gamma(q_t)(\mathcal{G}_{\eta})-\xi_t,
\end{equation}
where $\xi_t\geq 0$ captures approximation error from retrieval, feedback, LLM generation, and finite rollout estimates. 
Let $p_t=q_t(\mathcal{G}_{\eta})$ and $r=q^\star(\mathcal{G}_{\eta})$. 
Then
\begin{equation}
    q_{t+1}(\mathcal{G}_{\eta})
    \geq
    \frac{p_t+\gamma r}{1+\gamma}-\xi_t.
\end{equation}
Hence $q_{t+1}(\mathcal{G}_{\eta})>p_t$ whenever
\begin{equation}
    \xi_t
    <
    \frac{\gamma}{1+\gamma}(r-p_t).
\end{equation}
This condition is sufficient, not automatic: if the correction signal is too noisy or not more concentrated on $\mathcal{G}_{\eta}$ than the current proposal, the theory does not claim improvement.

\subsection{Comparison with a Generic LLM Baseline}

Consider WMAttack and a generic LLM-based search baseline under the same evaluator, task, search space, and budget. 
Let their effective correction strengths be $\gamma_{\mathrm{ours}}$ and $\gamma_{\mathrm{base}}$, respectively. 
Assume both methods start from the same proposal mass $p_t=q_t(\mathcal{G}_{\eta})$ and are compared against the same reference mass $r=q^\star(\mathcal{G}_{\eta})$.

The difference between their ideal corrected masses is
\begin{align}
    C_{\gamma_{\mathrm{ours}}}(q_t)(\mathcal{G}_{\eta})
    -
    C_{\gamma_{\mathrm{base}}}(q_t)(\mathcal{G}_{\eta})
    &=
    \frac{p_t+\gamma_{\mathrm{ours}}r}{1+\gamma_{\mathrm{ours}}}
    -
    \frac{p_t+\gamma_{\mathrm{base}}r}{1+\gamma_{\mathrm{base}}} \notag\\
    &=
    \frac{
    (\gamma_{\mathrm{ours}}-\gamma_{\mathrm{base}})(r-p_t)
    }{
    (1+\gamma_{\mathrm{ours}})(1+\gamma_{\mathrm{base}})
    }.
\end{align}
If the realized proposal updates have approximation errors $\xi_{\mathrm{ours}}$ and $\xi_{\mathrm{base}}$, then WMAttack assigns more mass to $\mathcal{G}_{\eta}$ than the baseline whenever
\begin{equation}
    \xi_{\mathrm{ours}}+\xi_{\mathrm{base}}
    <
    \frac{
    (\gamma_{\mathrm{ours}}-\gamma_{\mathrm{base}})(r-p_t)
    }{
    (1+\gamma_{\mathrm{ours}})(1+\gamma_{\mathrm{base}})
    },
\end{equation}
provided $\gamma_{\mathrm{ours}}>\gamma_{\mathrm{base}}$ and $r>p_t$. 
This formalizes a sufficient condition under which representation-guided retrieval and feedback improve the one-round probability of hitting an effective attack compared with a generic LLM proposal mechanism.

\subsection{Finite-Episode Utility Estimation}

The main analysis uses the population utility $U_\tau(c)$, while experiments observe an empirical estimate $\widehat{U}_\tau(c)$ from a finite number of rollout episodes. 
Assume episodic returns lie in $[R_{\min},R_{\max}]$ and action-flip indicators are Bernoulli. 
Ignoring deterministic runtime terms and absorbing variability-estimation error into the same bound, Hoeffding's inequality and a union bound over $\mathcal{C}_\tau$ imply that, with probability at least $1-\delta$,
\begin{equation}
    \left|
    \widehat{U}_\tau(c)-U_\tau(c)
    \right|
    \leq
    \zeta_m(\delta)
    \qquad
    \text{for all } c\in\mathcal{C}_\tau,
\end{equation}
where
\begin{equation}
    \zeta_m(\delta)
    =
    \frac{R_{\max}-R_{\min}}
    {|J_{\mathrm{clean}}(\tau)|+1}
    \sqrt{\frac{\log(4|\mathcal{C}_\tau|/\delta)}{2m}}
    +
    w_f
    \sqrt{\frac{\log(4|\mathcal{C}_\tau|/\delta)}{2m}}.
\end{equation}
Consequently, if a configuration is empirically $\eta$-optimal,
\begin{equation}
    \widehat{U}_\tau(c)
    \geq
    \max_{c'\in\mathcal{C}_\tau}
    \widehat{U}_\tau(c')-\eta,
\end{equation}
then it is population $(\eta+2\zeta_m(\delta))$-optimal:
\begin{equation}
    U_\tau(c)
    \geq
    U_\tau^\star-\eta-2\zeta_m(\delta).
\end{equation}
Thus, finite-episode evaluation enlarges the effective-set tolerance by a statistical estimation term rather than invalidating the search analysis.
\clearpage
\section{Task-Level Main Results}
\label{app:attack-family-selection}

This appendix expands the main results with task-level statistics.
Unless otherwise specified, the Atari WMAttack rows use the Qwen2.5-3B backend selected in the main text, while the DMC rows use the main DreamerV3 DMC run.
Rows marked with ``--'' indicate that the corresponding attack family was not available in the current export for that suite.

\begin{table}[htbp]
\centering
\caption{Attack-family selection counts for the strongest WMAttack configuration on each task.For Atari, Square Attack is listed for completeness but is excluded from the primary Atari runs due to its closed-loop evaluation cost.}
\label{tab:app-attack-family-counts}
\small
\begin{tabular}{lrrrrr}
\toprule
Suite & APGD-CE & APGD-DLR & FAB & Square & PhysCond-WMA \\
\midrule
Atari & 17 & 3 & 2 & 0 & 4 \\
DMC & 3 & 1 & 14 & 0 & 2 \\
\bottomrule
\end{tabular}
\end{table}

\begin{table}[htbp]
\centering
\caption{Task-level comparison on DreamerV3 Atari.}
\label{tab:app-atari-task-comparison}
\scriptsize
\begin{tabular}{lrrrrrr}
\toprule
Task & Claudini D & Random D & WMAttack D & Claudini U & Random U & WMAttack U \\
\midrule
Alien & 0.380 & 0.447 & 0.671 & -0.142 & -0.115 & 0.195 \\
Amidar & 0.296 & 0.229 & 0.899 & -0.382 & -0.408 & 0.366 \\
Assault & 0.036 & 0.090 & 0.519 & -0.587 & -0.560 & -0.098 \\
Asterix & 0.603 & 0.397 & 0.775 & 0.088 & -0.105 & 0.370 \\
BankHeist & 0.748 & 0.797 & 0.974 & 0.074 & 0.121 & 0.459 \\
BattleZone & 0.473 & 0.371 & 0.710 & -0.146 & -0.257 & 0.106 \\
Boxing & 0.630 & 0.698 & 0.973 & 0.021 & 0.056 & 0.332 \\
Breakout & 0.927 & 0.906 & 0.879 & 0.565 & 0.538 & 0.370 \\
ChopperCommand & 0.421 & 0.207 & 0.842 & -0.170 & -0.392 & 0.330 \\
CrazyClimber & 0.126 & 0.302 & 0.755 & -0.690 & -0.443 & 0.032 \\
DemonAttack & 0.362 & 0.256 & 0.740 & -0.188 & -0.338 & 0.002 \\
Freeway & 0.000 & 0.000 & 0.000 & -0.628 & -0.614 & -0.621 \\
Frostbite & 0.246 & 0.202 & 0.751 & -0.285 & -0.305 & 0.068 \\
Gopher & 0.531 & 0.235 & 0.846 & -0.199 & -0.495 & 0.245 \\
Hero & 0.014 & 0.096 & 0.643 & -0.554 & -0.484 & 0.229 \\
Jamesbond & 0.185 & 0.370 & 0.840 & -0.390 & -0.189 & 0.076 \\
Kangaroo & 0.549 & 0.477 & 0.892 & -0.166 & -0.199 & 0.317 \\
Krull & 0.169 & 0.156 & 0.611 & -0.560 & -0.554 & -0.090 \\
KungFuMaster & 0.440 & 0.311 & 0.903 & -0.269 & -0.355 & 0.072 \\
MsPacman & 0.388 & 0.439 & 0.697 & -0.211 & -0.055 & 0.193 \\
Pong & 1.942 & 1.935 & 1.971 & 1.401 & 1.403 & 1.453 \\
PrivateEye & 2.460 & 1.287 & 7.146 & 1.571 & 0.459 & 5.984 \\
Qbert & 0.197 & 0.771 & 0.966 & -0.526 & 0.188 & 0.505 \\
RoadRunner & 0.354 & 0.255 & 0.887 & -0.264 & -0.399 & 0.396 \\
Seaquest & 0.220 & 0.137 & 0.275 & -0.404 & -0.418 & -0.295 \\
UpNDown & 0.217 & 0.233 & 0.709 & -0.394 & -0.326 & 0.196 \\
\bottomrule
\end{tabular}
\end{table}

\begin{table}[htbp]
\centering
\caption{Task-level comparison on DreamerV3 DMC.}
\label{tab:app-dmc-task-comparison}
\scriptsize
\begin{tabular}{lrrrrrr}
\toprule
Task & Claudini D & Random D & WMAttack D & Claudini U & Random U & WMAttack U \\
\midrule
acrobot swingup & 0.472 & 0.657 & 0.986 & -0.175 & -0.003 & 0.235 \\
ball in cup catch & 0.074 & 0.049 & 0.865 & -0.366 & -0.368 & 0.264 \\
cartpole balance & 0.273 & 0.319 & 0.510 & -0.158 & -0.153 & 0.071 \\
cartpole balance sparse & 0.277 & 0.647 & 0.953 & -0.265 & -0.074 & 0.356 \\
cartpole swingup & 0.202 & 0.273 & 0.755 & -0.282 & -0.266 & 0.180 \\
cartpole swingup sparse & 0.756 & 0.632 & 0.988 & 0.220 & 0.018 & 0.455 \\
cheetah run & 0.620 & 0.631 & 0.845 & 0.225 & 0.225 & 0.445 \\
finger spin & 0.545 & 0.425 & 0.882 & -0.128 & -0.232 & 0.096 \\
finger turn easy & 0.197 & 0.177 & 0.553 & -0.166 & -0.209 & 0.110 \\
finger turn hard & 0.171 & 0.209 & 0.472 & -0.359 & -0.265 & 0.021 \\
hopper hop & 0.854 & 0.839 & 0.950 & 0.457 & 0.437 & 0.590 \\
hopper stand & 0.898 & 0.996 & 0.993 & 0.334 & 0.514 & 0.468 \\
pendulum swingup & 0.000 & 0.000 & 0.000 & -0.517 & -0.547 & -0.629 \\
quadruped run & 0.346 & 0.339 & 0.643 & 0.012 & -0.009 & 0.307 \\
quadruped walk & 0.217 & 0.386 & 0.884 & -0.141 & -0.016 & 0.184 \\
reacher easy & 0.009 & 0.037 & 0.139 & -0.357 & -0.351 & -0.285 \\
reacher hard & 0.093 & 0.064 & 0.860 & -0.286 & -0.342 & 0.424 \\
walker run & 0.287 & 0.307 & 0.772 & -0.116 & -0.097 & 0.384 \\
walker stand & 0.027 & 0.022 & 0.149 & -0.338 & -0.383 & -0.272 \\
walker walk & 0.057 & 0.034 & 0.447 & -0.332 & -0.343 & 0.069 \\
\bottomrule
\end{tabular}
\end{table}

\begin{table}[htbp]
\centering
\caption{Best discovered WMAttack configuration for each DreamerV3 Atari task.}
\label{tab:app-atari-best-config}
\scriptsize
\begin{tabular}{llrrlrrr}
\toprule
Task & Attack & $\epsilon$ & Steps & Alloc. & Drop $\uparrow$ & Flip $\uparrow$ & Utility $\uparrow$ \\
\midrule
Alien & APGD-CE & 20 & 24 & margin-linear & 0.671 & 0.684 & 0.195 \\
Amidar & APGD-CE & 20 & 24 & fixed & 0.899 & 0.686 & 0.366 \\
Assault & APGD-CE & 8 & 10 & margin-linear & 0.519 & 0.521 & -0.098 \\
Asterix & APGD-DLR & 8 & 6 & margin-linear & 0.775 & 0.623 & 0.370 \\
BankHeist & APGD-CE & 20 & 20 & margin-linear & 0.974 & 0.743 & 0.459 \\
BattleZone & APGD-CE & 8 & 10 & margin-linear & 0.710 & 0.704 & 0.106 \\
Boxing & APGD-CE & 20 & 20 & fixed & 0.973 & 0.782 & 0.332 \\
Breakout & APGD-DLR & 10 & 12 & margin-linear & 0.879 & 0.476 & 0.370 \\
ChopperCommand & APGD-CE & 20 & 4 & fixed & 0.842 & 0.570 & 0.330 \\
CrazyClimber & APGD-CE & 8 & 6 & margin-linear & 0.755 & 0.706 & 0.032 \\
DemonAttack & PhysCond-WMA & 22 & 18 & margin-linear & 0.740 & 0.815 & 0.002 \\
Freeway & FAB & 10 & 18 & margin-linear & 0.000 & 0.524 & -0.621 \\
Frostbite & PhysCond-WMA & 22 & 24 & margin-linear & 0.751 & 0.838 & 0.068 \\
Gopher & APGD-CE & 8 & 6 & margin-linear & 0.846 & 0.592 & 0.245 \\
Hero & APGD-CE & 20 & 24 & margin-linear & 0.643 & 0.761 & 0.229 \\
Jamesbond & PhysCond-WMA & 22 & 18 & margin-linear & 0.840 & 0.781 & 0.076 \\
Kangaroo & APGD-CE & 22 & 22 & margin-linear & 0.892 & 0.570 & 0.317 \\
Krull & APGD-CE & 20 & 24 & margin-linear & 0.611 & 0.664 & -0.090 \\
KungFuMaster & PhysCond-WMA & 26 & 14 & margin-linear & 0.903 & 0.837 & 0.072 \\
MsPacman & APGD-CE & 20 & 10 & margin-linear & 0.697 & 0.634 & 0.193 \\
Pong & FAB & 10 & 12 & margin-linear & 1.971 & 0.322 & 1.453 \\
PrivateEye & APGD-CE & 10 & 12 & margin-linear & 7.146 & 0.702 & 5.984 \\
Qbert & APGD-DLR & 8 & 6 & margin-linear & 0.966 & 0.567 & 0.505 \\
RoadRunner & APGD-CE & 22 & 22 & margin-linear & 0.887 & 0.703 & 0.396 \\
Seaquest & APGD-CE & 10 & 16 & margin-linear & 0.275 & 0.646 & -0.295 \\
UpNDown & APGD-CE & 8 & 6 & margin-linear & 0.709 & 0.720 & 0.196 \\
\bottomrule
\end{tabular}
\end{table}

\begin{table}[htbp]
\centering
\caption{Best discovered WMAttack configuration for each DreamerV3 DMC task.}
\label{tab:app-dmc-best-config}
\scriptsize
\begin{tabular}{llrrlrrr}
\toprule
Task & Attack & $\epsilon$ & Steps & Alloc. & Drop $\uparrow$ & Flip $\uparrow$ & Utility $\uparrow$ \\
\midrule
acrobot swingup & APGD-DLR & 10 & 14 & margin-linear & 0.986 & 0.328 & 0.235 \\
ball in cup catch & FAB & 10 & 18 & fixed & 0.865 & 0.657 & 0.264 \\
cartpole balance & FAB & 10 & 20 & fixed & 0.510 & 0.825 & 0.071 \\
cartpole balance sparse & APGD-CE & 10 & 10 & margin-linear & 0.953 & 0.548 & 0.356 \\
cartpole swingup & FAB & 10 & 20 & fixed & 0.755 & 0.718 & 0.180 \\
cartpole swingup sparse & APGD-CE & 10 & 12 & margin-linear & 0.988 & 0.704 & 0.455 \\
cheetah run & FAB & 10 & 16 & fixed & 0.845 & 0.960 & 0.445 \\
finger spin & PhysCond-WMA & 10 & 10 & margin-linear & 0.882 & 0.511 & 0.096 \\
finger turn easy & FAB & 10 & 18 & fixed & 0.553 & 0.937 & 0.110 \\
finger turn hard & FAB & 10 & 20 & fixed & 0.472 & 0.932 & 0.021 \\
hopper hop & FAB & 10 & 18 & fixed & 0.950 & 0.950 & 0.590 \\
hopper stand & FAB & 10 & 18 & margin-linear & 0.993 & 0.830 & 0.468 \\
pendulum swingup & APGD-CE & 10 & 10 & margin-linear & 0.000 & 0.457 & -0.629 \\
quadruped run & FAB & 10 & 14 & fixed & 0.643 & 0.994 & 0.307 \\
quadruped walk & PhysCond-WMA & 10 & 18 & margin-linear & 0.884 & 1.000 & 0.184 \\
reacher easy & FAB & 10 & 20 & fixed & 0.139 & 0.943 & -0.285 \\
reacher hard & FAB & 10 & 16 & fixed & 0.860 & 0.904 & 0.424 \\
walker run & FAB & 10 & 16 & fixed & 0.772 & 0.942 & 0.384 \\
walker stand & FAB & 10 & 20 & fixed & 0.149 & 0.919 & -0.272 \\
walker walk & FAB & 10 & 14 & fixed & 0.447 & 0.949 & 0.069 \\
\bottomrule
\end{tabular}
\end{table}

\begin{table}[htbp]
\centering
\caption{Preliminary DreamerV2 Atari task-level results.}
\label{tab:app-dv2-atari-best-config}
\scriptsize
\begin{tabular}{llrrlrrr}
\toprule
Task & Attack & $\epsilon$ & Steps & Alloc. & Drop $\uparrow$ & Flip $\uparrow$ & Utility $\uparrow$ \\
\midrule
Alien & APGD-CE & 10 & 14 & margin-linear & 0.312 & 0.650 & -0.185 \\
Amidar & APGD-CE & 10 & 14 & margin-linear & 0.241 & 0.766 & -0.222 \\
Assault & APGD-CE & 10 & 10 & margin-linear & 0.253 & 0.386 & -0.386 \\
Asterix & APGD-CE & 10 & 14 & margin-linear & 0.452 & 0.675 & 0.118 \\
BankHeist & APGD-CE & 10 & 8 & margin-linear & 0.950 & 0.580 & 0.365 \\
\bottomrule
\end{tabular}
\end{table}

\begin{table}[htbp]
\centering
\caption{Per-task, per-attack-family results on DreamerV3 Atari. Each cell reports Drop / Utility for the best confirmed configuration within an attack family.}
\label{tab:app-atari-task-attack-matrix}
\scriptsize
\begin{tabular}{lcccc}
\toprule
Task & APGD-CE & APGD-DLR & FAB  & PhysCond-WMA \\
\midrule
Alien & 0.671 / 0.195 & 0.611 / 0.048 & 0.170 / -0.344  & 0.722 / -0.101 \\
Amidar & 0.899 / 0.366 & 0.722 / 0.094 & 0.231 / -0.366  & 0.962 / 0.177 \\
Assault & 0.519 / -0.098 & 0.525 / -0.150 & 0.213 / -0.388  & 0.501 / -0.267 \\
Asterix & 0.744 / 0.354 & 0.775 / 0.370 & 0.214 / -0.257  & 0.775 / 0.164 \\
BankHeist & 0.974 / 0.459 & 0.973 / 0.423 & 0.088 / -0.600  & 0.923 / 0.250 \\
BattleZone & 0.710 / 0.106 & 0.579 / -0.075 & 0.526 / -0.097  & 0.803 / -0.052 \\
Boxing & 0.973 / 0.332 & 0.929 / 0.249 & 0.111 / -0.384  & 1.002 / 0.044 \\
Breakout & 0.739 / 0.083 & 0.879 / 0.370 & 0.364 / -0.167  & 0.933 / 0.300 \\
ChopperCommand & 0.842 / 0.330 & 0.513 / -0.150 & 0.302 / -0.260  & 0.736 / -0.004 \\
CrazyClimber & 0.755 / 0.032 & 0.807 / 0.017 & 0.093 / -0.719  & 0.773 / -0.078 \\
DemonAttack & 0.633 / -0.041 & 0.391 / -0.255 & 0.453 / -0.147 & 0.740 / 0.002 \\
Freeway & 0.000 / -0.648 & 0.000 / -0.695 & 0.000 / -0.621  & 0.000 / -0.899 \\
Frostbite & 0.190 / -0.410 & 0.413 / -0.123 & 0.128 / -0.390  & 0.751 / 0.068 \\
Gopher & 0.846 / 0.245 & 0.789 / 0.093 & 0.410 / -0.270  & 0.912 / 0.138 \\
Hero & 0.643 / 0.229 & 0.248 / -0.330 & 0.011 / -0.522  & 0.772 / 0.157 \\
Jamesbond & 0.341 / -0.444 & 0.656 / -0.021 & 0.184 / -0.392  & 0.840 / 0.076 \\
Kangaroo & 0.892 / 0.317 & 0.883 / 0.262 & 0.468 / -0.148  & 0.865 / 0.136 \\
Krull & 0.611 / -0.090 & 0.308 / -0.434 & 0.165 / -0.490  & 0.449 / -0.372 \\
KungFuMaster & 0.596 / -0.071 & 0.493 / -0.196 & -0.113 / -0.806  & 0.903 / 0.072 \\
MsPacman & 0.697 / 0.193 & 0.730 / 0.190 & 0.276 / -0.295  & 0.837 / 0.170 \\
Pong & 1.854 / 1.128 & 1.961 / 1.245 & 1.971 / 1.453  & 1.942 / 1.126 \\
PrivateEye & 7.146 / 5.984 & 1.423 / 0.613 & -3.286 / -3.997  & 2.676 / 1.710 \\
Qbert & 0.854 / 0.176 & 0.966 / 0.505 & -0.026 / -0.637  & 0.971 / 0.402 \\
RoadRunner & 0.887 / 0.396 & 0.550 / -0.106 & 0.275 / -0.384  & 0.607 / -0.104 \\
Seaquest & 0.275 / -0.295 & 0.241 / -0.370 & 0.023 / -0.550  & 0.332 / -0.515 \\
UpNDown & 0.709 / 0.196 & 0.676 / 0.129 & 0.233 / -0.285  & 0.890 / 0.084 \\
\bottomrule
\end{tabular}
\end{table}

\begin{table}[htbp]
\centering
\caption{Per-task, per-attack-family results on DreamerV3 DMC. Each cell reports Drop / Utility for the best confirmed configuration within an attack family.}
\label{tab:app-dmc-task-attack-matrix}
\scriptsize
\begin{tabular}{lccccc}
\toprule
Task & APGD-CE & APGD-DLR & FAB & Square & PhysCond-WMA \\
\midrule
acrobot swingup & 0.997 / 0.180 & 0.986 / 0.235 & 0.991 / 0.178 & 0.201 / -0.655 & 0.872 / -0.134 \\
ball in cup catch & 0.235 / -0.280 & 0.411 / -0.132 & 0.865 / 0.264 & 0.115 / -0.447 & 0.817 / 0.010 \\
cartpole balance & 0.385 / -0.124 & 0.392 / -0.071 & 0.510 / 0.071 & 0.001 / -0.915 & 0.411 / -0.305 \\
cartpole balance sparse & 0.953 / 0.356 & 0.839 / 0.277 & 0.958 / 0.342 & 0.002 / -0.755 & 0.922 / 0.093 \\
cartpole swingup & 0.640 / 0.089 & 0.628 / 0.100 & 0.755 / 0.180 & 0.068 / -0.649 & 0.659 / -0.109 \\
cartpole swingup sparse & 0.988 / 0.455 & 0.974 / 0.389 & 0.994 / 0.362 & 0.028 / -0.643 & 0.960 / 0.159 \\
cheetah run & 0.535 / 0.083 & 0.575 / 0.141 & 0.845 / 0.445 & 0.386 / -0.152 & 0.843 / 0.147 \\
finger spin & 0.420 / -0.218 & 0.505 / -0.159 & 0.614 / -0.122 & 0.730 / -0.056 & 0.882 / 0.096 \\
finger turn easy & 0.233 / -0.176 & 0.276 / -0.124 & 0.553 / 0.110 & 0.178 / -0.279 & 0.339 / -0.372 \\
finger turn hard & 0.111 / -0.439 & 0.294 / -0.180 & 0.472 / 0.021 & -0.124 / -0.771 & 0.343 / -0.385 \\
hopper hop & 0.799 / 0.342 & 0.831 / 0.373 & 0.950 / 0.590 & 0.224 / -0.491 & 0.971 / 0.289 \\
hopper stand & 0.948 / 0.150 & 0.921 / 0.407 & 0.993 / 0.468 & -0.157 / -0.668 & 0.993 / 0.095 \\
pendulum swingup & 0.000 / -0.629 & 0.000 / -0.654 & 0.000 / -0.657 & 0.000 / -0.815 & 0.000 / -0.839 \\
quadruped run & -0.022 / -0.380 & -0.058 / -0.397 & 0.643 / 0.307 & -0.172 / -0.550 & 0.925 / 0.228 \\
quadruped walk & 0.223 / -0.152 & 0.177 / -0.195 & 0.494 / 0.160 & 0.170 / -0.214 & 0.884 / 0.184 \\
reacher easy & 0.034 / -0.391 & 0.111 / -0.308 & 0.139 / -0.285 & 0.036 / -0.407 & 0.236 / -0.518 \\
reacher hard & 0.325 / -0.113 & 0.317 / -0.093 & 0.860 / 0.424 & 0.093 / -0.350 & 0.616 / -0.098 \\
walker run & 0.556 / 0.136 & 0.549 / 0.139 & 0.772 / 0.384 & 0.018 / -0.623 & 0.762 / 0.044 \\
walker stand & 0.068 / -0.394 & -0.032 / -0.409 & 0.149 / -0.272 & -0.043 / -0.506 & 0.192 / -0.532 \\
walker walk & 0.162 / -0.241 & 0.269 / -0.145 & 0.447 / 0.069 & -0.013 / -0.475 & 0.321 / -0.385 \\
\bottomrule
\end{tabular}
\end{table}
\clearpage
\section{Efficiency-Curve Details}
\label{app:efficiency-details}

This section provides the numerical values behind the attack-specific efficiency curves in Figure~\ref{fig:efficiency-curves}. 
For each suite, attack family, and method, we report the mean best-so-far scalarized utility after $1$, $2$, $4$, and $8$ evaluated configurations. 
The averages are computed across tasks. 
The shaded regions in Figure~\ref{fig:efficiency-curves} correspond to standard errors across tasks; here we report only the mean values for compactness.

\begin{table}[htbp]
\centering
\caption{Attack-specific search efficiency on DreamerV3 Atari. Each value is the mean best-so-far scalarized utility after a given number of evaluated configurations.}
\label{tab:app-efficiency-atari}
\small
\begin{tabular}{llrrrr}
\toprule
Attack & Method & Trial 1 & Trial 2 & Trial 4 & Trial 8 \\
\midrule
APGD-CE & Claudini & -0.175 & -0.102 & 0.032 & 0.193 \\
APGD-CE & Random & -0.161 & -0.080 & 0.124 & 0.202 \\
APGD-CE & WMAttack & \textbf{0.183} & \textbf{0.522} & \textbf{0.582} & \textbf{0.651} \\
\midrule
APGD-DLR & Claudini & -0.158 & -0.067 & 0.010 & 0.095 \\
APGD-DLR & Random & -0.155 & -0.114 & -0.068 & 0.029 \\
APGD-DLR & WMAttack & \textbf{0.269} & \textbf{0.286} & \textbf{0.409} & \textbf{0.516} \\
\midrule
FAB & Claudini & -0.609 & -0.401 & -0.258 & 0.006 \\
FAB & Random & -0.476 & -0.438 & \textbf{-0.111} & \textbf{0.069} \\
FAB & WMAttack & \textbf{-0.355} & \textbf{-0.332} & -0.236 & -0.070 \\
\midrule
PhysCond-WMA & Claudini & -0.224 & -0.182 & 0.171 & 0.259 \\
PhysCond-WMA & Random & -0.260 & -0.200 & -0.099 & 0.010 \\
PhysCond-WMA & WMAttack & \textbf{0.796} & \textbf{0.801} & \textbf{0.925} & \textbf{1.015} \\
\bottomrule
\end{tabular}
\end{table}

\begin{table}[htbp]
\centering
\caption{Attack-specific search efficiency on DreamerV3 DMC. Each value is the mean best-so-far scalarized utility after a given number of evaluated configurations.}
\label{tab:app-efficiency-dmc}
\small
\begin{tabular}{llrrrr}
\toprule
Attack & Method & Trial 1 & Trial 2 & Trial 4 & Trial 8 \\
\midrule
APGD-CE & Claudini & -0.303 & -0.241 & -0.196 & -0.158 \\
APGD-CE & Random & -0.273 & -0.238 & -0.175 & -0.131 \\
APGD-CE & WMAttack & \textbf{-0.039} & \textbf{-0.011} & \textbf{0.024} & \textbf{0.037} \\
\midrule
APGD-DLR & Claudini & -0.365 & -0.282 & -0.196 & -0.168 \\
APGD-DLR & Random & -0.226 & -0.206 & -0.170 & -0.151 \\
APGD-DLR & WMAttack & \textbf{-0.062} & \textbf{-0.045} & \textbf{0.005} & \textbf{0.065} \\
\midrule
Square & Claudini & -0.545 & -0.476 & -0.448 & -0.400 \\
Square & Random & \textbf{-0.526} & \textbf{-0.472} & \textbf{-0.437} & \textbf{-0.375} \\
Square & WMAttack & -0.621 & -0.577 & -0.515 & -0.440 \\
\midrule
PhysCond-WMA & Claudini & -0.410 & -0.377 & -0.326 & -0.275 \\
PhysCond-WMA & Random & -0.384 & -0.338 & -0.320 & -0.297 \\
PhysCond-WMA & WMAttack & \textbf{-0.106} & \textbf{-0.094} & \textbf{-0.069} & \textbf{-0.016} \\
\bottomrule
\end{tabular}
\end{table}

The Atari results show that WMAttack improves early-stage search quality for APGD-CE, APGD-DLR, and PhysCond-WMA, while FAB is more competitive. 
On DMC, WMAttack improves APGD-CE, APGD-DLR, and PhysCond-WMA, whereas Square Attack remains difficult for all methods and is better optimized by the baselines in this export. 
These results support the main observation that retrieval and feedback improve finite-budget search for most attack families, while the size of the gain depends on the attack family and task suite.
\clearpage
\section{Ablation Details}
\label{app:ablation-details}

This section provides additional statistics for the RGAR ablation in Section~5.4. 
The experiment is conducted on DreamerV3 Atari with 26 tasks and four attack families: APGD-CE, APGD-DLR, FAB, and PhysCond-WMA. 
We compare RGAR initialization with random initialization while keeping the downstream SCAS protocol fixed. 
Each task--attack pair is treated as one comparison unit, resulting in 104 pairs.
\begin{figure}[t]
    \centering
    \begin{minipage}[t]{0.49\linewidth}
        \centering
        \includegraphics[width=\linewidth]{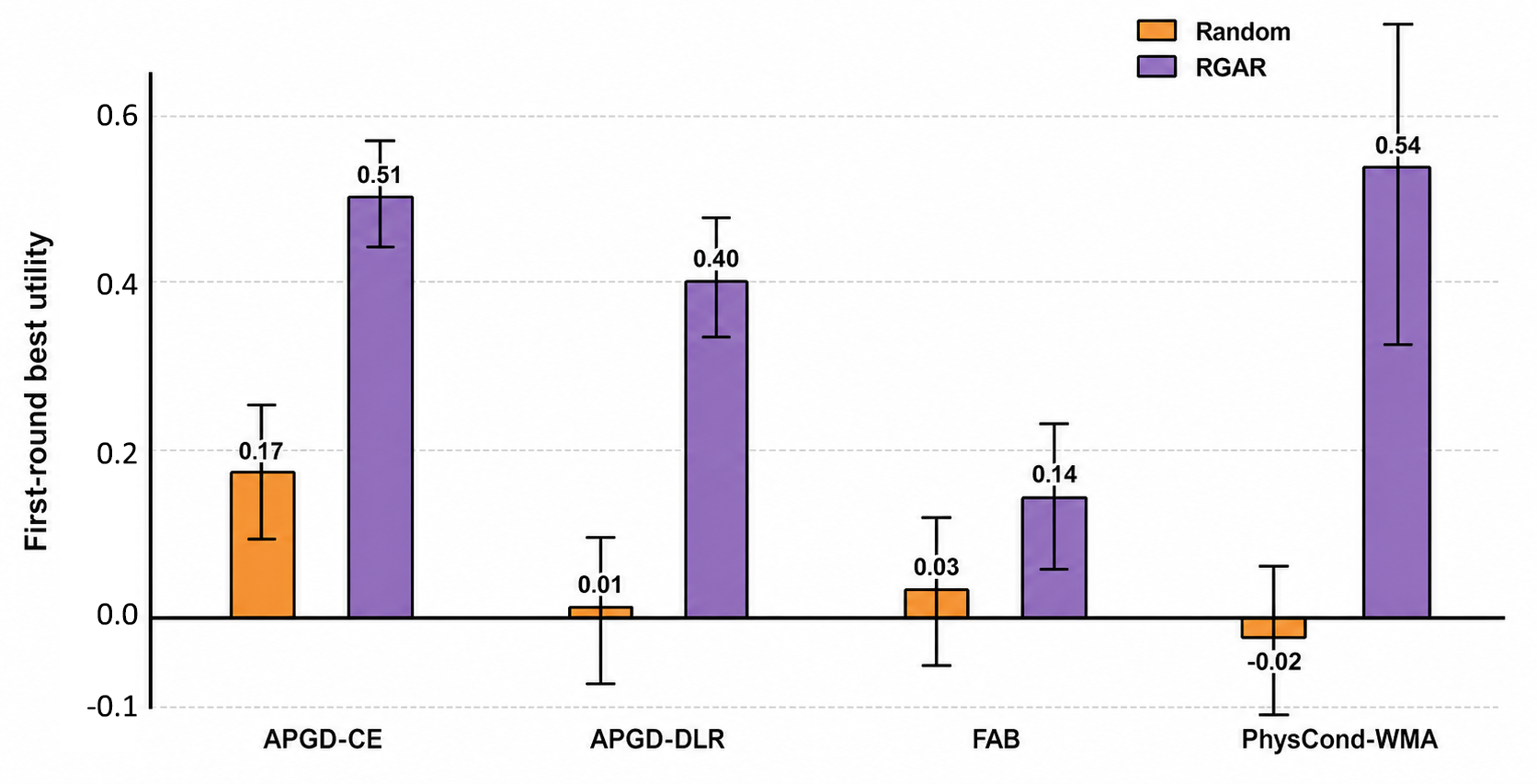}
        \vspace{1mm}
        \centerline{\small (a) RGAR warm-start effect}
    \end{minipage}
    \hfill
    \begin{minipage}[t]{0.49\linewidth}
        \centering
        \includegraphics[width=\linewidth]{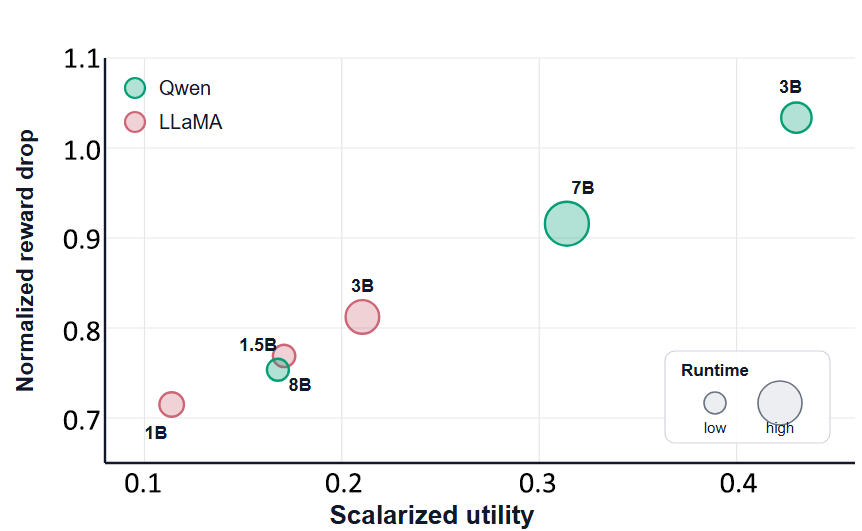}
        \vspace{1mm}
        \centerline{\small (b) LLM backbone sensitivity}
    \end{minipage}
    \vspace{1mm}
    \caption{
    Design analysis of WMAttack on DreamerV3 Atari.
    RGAR improves first-round candidate quality, while LLM backbones induce different efficiency--accuracy trade-offs.
    }
    \label{fig:design-analysis}
\end{figure}
\begin{table}[htbp]
\centering
\caption{Per-attack RGAR ablation on DreamerV3 Atari. Each cell reports Utility / Drop / Flip averaged across 26 tasks.}
\label{tab:app-rgar-per-attack}
\small
\begin{tabular}{llcc}
\toprule
Attack & Variant & First Utility / Drop / Flip $\uparrow$ & Final Utility / Drop / Flip $\uparrow$ \\
\midrule
APGD-CE & Random init. 
& 0.173 / 0.522 / 0.443 
& -0.243 / 0.395 / 0.439 \\
APGD-CE & RGAR init. 
& \textbf{0.505 / 0.813 / 0.658} 
& \textbf{0.206 / 0.795 / 0.650} \\
\midrule
APGD-DLR & Random init. 
& 0.010 / 0.412 / 0.396 
& -0.354 / 0.322 / 0.393 \\
APGD-DLR & RGAR init. 
& \textbf{0.402 / 0.769 / 0.614} 
& \textbf{0.086 / 0.714 / 0.606} \\
\midrule
FAB & Random init. 
& 0.032 / 0.363 / 0.227 
& -0.315 / 0.288 / 0.237 \\
FAB & RGAR init. 
& \textbf{0.140 / 0.483 / 0.255} 
& \textbf{-0.284 / 0.331 / 0.251} \\
\midrule
PhysCond-WMA & Random init. 
& -0.023 / 0.483 / 0.509 
& -0.389 / 0.393 / 0.514 \\
PhysCond-WMA & RGAR init. 
& \textbf{0.542 / 1.017 / 0.767} 
& \textbf{0.074 / 0.819 / 0.764} \\
\bottomrule
\end{tabular}
\end{table}

\begin{table}[htbp]
\centering
\caption{Pair-level win counts for RGAR over random initialization on DreamerV3 Atari. Each entry reports the number of task--attack pairs where RGAR is strictly better than random initialization.}
\label{tab:app-rgar-pair-wins}
\small
\begin{tabular}{lccc}
\toprule
Metric & First round & Final result & Total pairs \\
\midrule
Utility & 85 & 87 & 104 \\
Reward drop & 84 & 85 & 104 \\
Flip rate & 96 & 92 & 104 \\
\bottomrule
\end{tabular}
\end{table}

\begin{table}[t]
\centering
\caption{Per-attack final pair-level win counts for RGAR over random initialization. Each entry reports the number of tasks where RGAR achieves a higher final metric than random initialization.}
\label{tab:app-rgar-pair-wins-by-attack}
\small
\begin{tabular}{lccc}
\toprule
Attack & Utility & Drop & Flip \\
\midrule
APGD-CE & 25 / 26 & 25 / 26 & 25 / 26 \\
APGD-DLR & 25 / 26 & 25 / 26 & 25 / 26 \\
FAB & 12 / 26 & 10 / 26 & 16 / 26 \\
PhysCond-WMA & 25 / 26 & 25 / 26 & 26 / 26 \\
\bottomrule
\end{tabular}
\end{table}

These results show that RGAR improves both the initial proposal quality and the final search outcome for most task--attack pairs. 
The improvement is strongest for APGD-CE, APGD-DLR, and PhysCond-WMA, while FAB is more competitive. 
This is consistent with Figure~\ref{fig:design-analysis}(a), where RGAR provides a stronger warm start across all four attack families but the magnitude of improvement depends on the attack family.



\end{document}